\documentclass[letterpaper, 10 pt, journal]{IEEEtran}

\IEEEoverridecommandlockouts                              
\usepackage{graphicx}
\usepackage[mathletters]{ucs}
\usepackage[utf8x]{inputenc}
\usepackage{booktabs}
\usepackage{comment}
\usepackage{graphics}
\usepackage[table,xcdraw]{xcolor}
\usepackage{multicol}
\usepackage{multirow}
\usepackage{algorithm}
\usepackage{lmodern,amsmath,amssymb}
\usepackage[noend]{algpseudocode}
\usepackage{caption}
\usepackage{paralist}
\usepackage{subcaption}
\usepackage{setspace}
\usepackage[hyphens]{url}
\algnewcommand\algorithmicforeach{\textbf{for each}}
\algdef{S}[FOR]{ForEach}[1]{\algorithmicforeach\ #1\ \algorithmicdo}
\title{\LARGE \bf
Neural Storage: A New Paradigm of Elastic Memory
}

\author{\large Prabuddha Chakraborty and Swarup Bhunia
\\  Department of Electrical \& Computer Engineering
\\University of Florida, Gainesville, FL, USA}

\usepackage[switch]{lineno}

\begin{document}
\nocite{comSurveyBhunia}

\maketitle
\thispagestyle{empty}
\pagestyle{empty}

\begin{abstract}
Storage and retrieval of data in a computer memory plays a major role in system performance. Traditionally, computer memory organization is ‘static’ – i.e., they do not change based on the application-specific characteristics in memory access behaviour during system operation. Specifically, the association of a data block with a search pattern (or cues) as well as the granularity of a stored data do not evolve. Such a static nature of computer memory, we observe, not only limits the amount of data we can store in a given physical storage, but it also misses the opportunity for dramatic performance improvement in various applications. On the contrary, human memory is characterized by seemingly infinite plasticity in storing and retrieving data – as well as dynamically creating/updating the associations between data and corresponding cues. In this paper, we introduce Neural Storage (NS), a brain-inspired learning memory paradigm that organizes the memory as a flexible neural memory network. In NS, the network structure, strength of associations, and granularity of the data adjust continuously during system operation, providing unprecedented plasticity and performance benefits. We present the associated storage/retrieval/retention algorithms in NS, which integrate a formalized learning process. Using a full-blown operational model, we demonstrate that NS achieves an order of magnitude improvement in memory access performance for two representative applications when compared to traditional content-based memory.

\end{abstract}


\section{Introduction}

Digital memory is an integral part of a computer system. It plays a major role in defining system performance. Memory access behaviour largely depends on the nature of the incoming data and the specific information-processing tasks that operate on the data. Applications ranging from wildlife surveillance \cite{wildlifeSurv} to infrastructure damage monitoring \cite{UAV_earthquake_IoT, UAVRoofHole_IoT} that collect, store and analyze data often exhibit distinct memory access (e.g., storage and retrieval of specific data blocks) behaviour. Even within the same application, such behaviour may change with time. Hence, these systems with variable and constantly evolving memory access pattern can benefit from a memory organization that can dynamically tailor itself to meet the requirements. Furthermore, many applications deal with multi-modal data (e.g., image and sound) \cite{multimodal, carthel2007multisensor_maritimeSurv} and in such applications, the data storage/access requires special considerations in terms of their temporal importance and inter-modality relations. A data storage framework which can efficiently store and retrieve multi-modal data is crucial for these applications. 

\begin{figure}[t]
\centering
\includegraphics[width=\columnwidth]{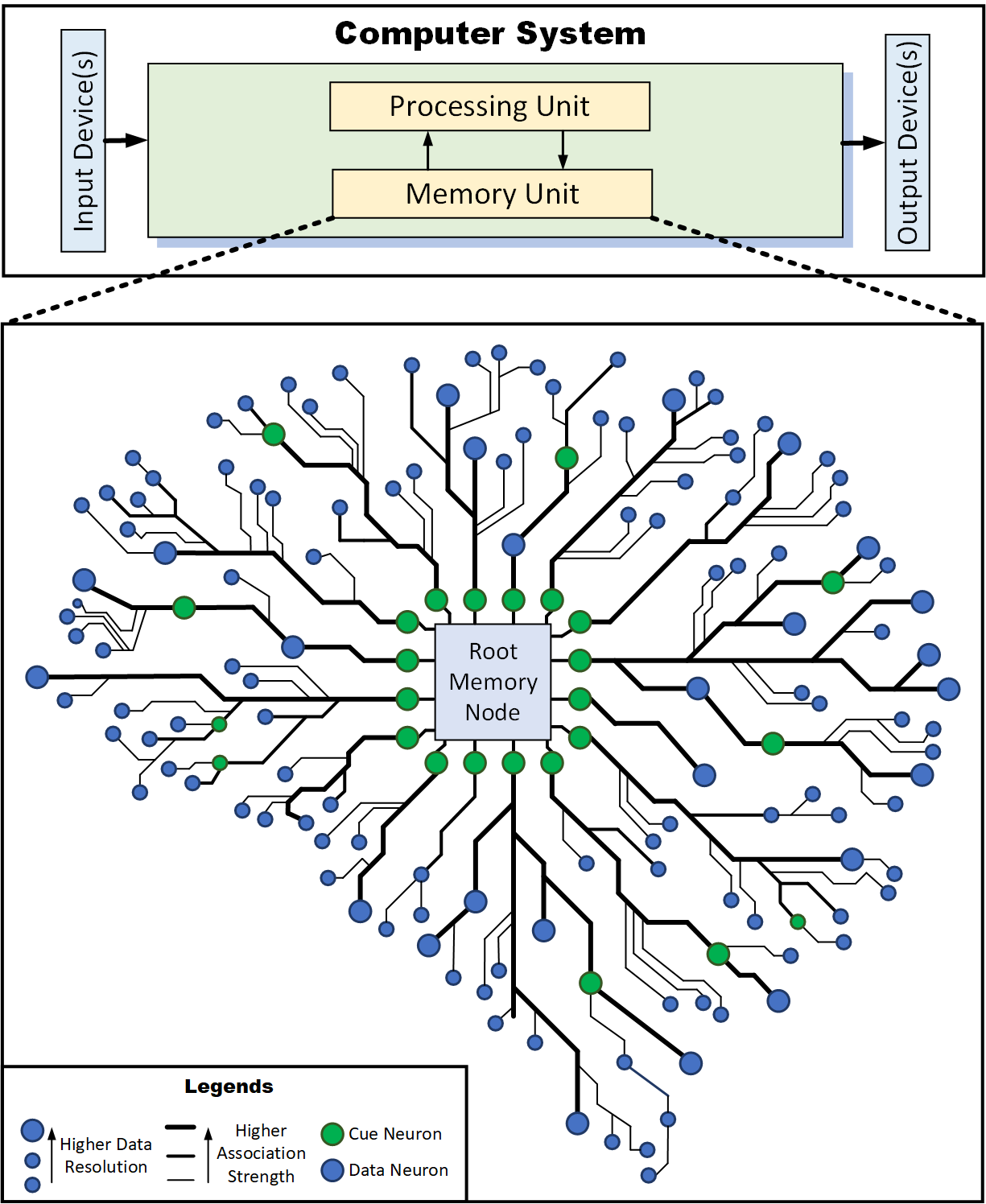}
\caption{Neural Storage: A new paradigm of elastic memory. }
\label{fig:teaser}
\vspace{-0.3in}
\end{figure}

Many computing systems, specifically the emergent internet of things (IoT) edge devices, come with tight constraints on memory storage capacity, energy and communication bandwidth \cite{AiIoT1000}. These systems often deal with a huge influx of data with varying degree of relevance to the application. 
Hence, storing and transmitting less useful data at higher quality may not be optimal. Due to these requirements, it is important for a memory framework to be efficient in terms of energy, space and transmission bandwidth utilization by focusing on what is important for the specific application.

Based on these observations, an ideal data storage framework for these applications should be:
\begin{itemize}
    \item Flexible and dynamic in nature to accommodate for the constantly evolving application requirements and scenarios.
    \item Able to emulate a virtually infinite memory that can deal with a huge influx of sensor data which is common in case of many IoT applications.
    \item Able to efficiently handle multi-modal data in the context of the application-specific requirements.
    \item Geared towards increasing storage, transmission and energy utilization efficiency.
\end{itemize}

Traditional memory frameworks \cite{comSurveyBhunia} (both address-operated and content-operated) are not ideal for meeting these requirements due to lack of flexibility in their memory organization and operations. In an address-operated memory, each address is associated with a data unit. And for a content-operated memory, each data-search-pattern (cue/tag) is associated with a single data unit. Hence, in both cases, the mapping is one-to-one and does not evolve without direct admin/user interference. Data in a traditional memory is also stored at a fixed quality/granularity. When a traditional memory runs out of space, it can either stop accepting new data or remove old data based on a specific data replacement policy. All these traits of a traditional memory are tied to its static nature which makes it not suitable for modern applications that have evolving needs and requirements as established earlier. For example, let us consider a wildlife image-based surveillance system which is geared towards detecting wolves. Any image frame that does not contain a wolf is considered to be of lower importance than any frame containing at least one wolf. However, a traditional memory, due to lack of dynamism in terms of data granularity management, will store the image frames at the same quality regardless of their importance to the application. Additionally, due to lack of dynamism in memory organization, searching for a wolf image will take the same effort/time as it would take for searching any rarely accessed and unimportant image.

To meet the requirements of many modern applications, it is attractive to incorporate flexibility and dynamism in the digital memory which we believe can be best achieved through statistics-guided learning. Artificial Intelligence (AI) and Machine Learning (ML) are widely used to solve different problems where static algorithms are not ideal. Similarly, meeting the dynamic memory requirements can not be possible using static algorithms. Hence incorporation of intelligence may be an ideal solution for addressing current digital memory limitations.

We draw inspiration from human biological memory which has many useful properties which can be beneficial for a digital memory as well. A human brain due to `plasticity' \cite{lindenberger2017towards_plasticity, lovden2010theoretical_plasticity}, undergoes internal change based on external stimuli and adapts to different scenarios presented to it. 
Data stored in a human brain is lossy in nature and are subject to decay and feature-loss. However, important memories decay at a slower rate and repetition/priming can lead to prolonged retention of important mission-critical data \cite{atkinson1968human_memoryDecay, tulving1990priming_primingHelps}. Human memory also receives and retains data from multiple sensory organs and intelligently stores this multi-modal data for optimal performance \cite{atkinson1968human_memoryDecay}. If these intelligence guided human memory properties can be realized in a digital memory with the help of ML, then it would be ideal for the emergent applications.

With this vision in mind, we put forward a paradigm-shifting content-operated memory framework, Neural Storage (NS) which mimics the intelligence of human brain for efficient storage and speedy access. In NS, the memory storage is a network of cues (search-patterns) and data, we term Neural Memory Network (NMN). Based on the feedback generated from each memory operation we use reinforcement learning to (1) optimize the NMN data/cue organization and (2) adjust the granularity (feature quality) of specific data units. NS is designed to have the same interface as any traditional Content Addressable Memory (CAM). This allows NS to efficiently replace traditional CAMs in any application as shown in Fig.~\ref{fig:teaser}. Applications which are resistant to imprecise data storage/retrieval and deals with storing data of varying importance will benefit the most from using NS. 

For quantitatively analyzing the effectiveness of using NS as a memory system, we implement a NS memory simulator with an array of tunable hyperparameter. We run different real-life applications using NS and observe that the NS framework utilizes orders of magnitude less space, and exhibits higher retrieval efficiency while incurring minimal impact on the application performance. 


\begin{table*}[]
\centering
\caption{Comparison between NS and traditional memory frameworks. The dynamic nature of NS, guided by continuous reinforcement learning, makes it adaptable to the application requirements and the usage scenarios.}
\label{memComp}
\renewcommand{\arraystretch}{1.3}
\scriptsize\addtolength{\tabcolsep}{-5pt}
\begin{tabular}{|c|c|c|c|c|c|c|c|}
\hline
\rowcolor[HTML]{EFEFEF} 
\cellcolor[HTML]{EFEFEF}                                 & \cellcolor[HTML]{EFEFEF}                                    & \multicolumn{2}{c|}{\cellcolor[HTML]{EFEFEF}\textbf{Dynamism}}                                                                                        & \multicolumn{3}{c|}{\cellcolor[HTML]{EFEFEF}\textbf{Associativity}}                                                                     & \cellcolor[HTML]{EFEFEF}                                                                                       \\ \cline{3-7}
\rowcolor[HTML]{EFEFEF} 
\multirow{-2}{*}{\cellcolor[HTML]{EFEFEF}\textbf{Memory Organization}}      & \multirow{-2}{*}{\cellcolor[HTML]{EFEFEF}\textbf{Learning}} & \textbf{\begin{tabular}[c]{@{}c@{}}Data\\ Resolution\end{tabular}}        & \textbf{Association}                                                      & \textbf{\textless{}Cue, Data\textgreater{}} & \textbf{\textless{}Data, Data\textgreater{}} & \textbf{\textless{}Cue, Cue\textgreater{}} & \multirow{-2}{*}{\cellcolor[HTML]{EFEFEF}\textbf{\begin{tabular}[c]{@{}c@{}}Space \\ Efficiency\end{tabular}}} \\ \hline
\cellcolor[HTML]{DAE8FC}\textbf{Address Operated Memory} & N/A                                                         & Fixed                                                                     & User defined                                                              & N/A                                         & N/A                                          & N/A                                        & Low                                                                                                            \\ \hline
\cellcolor[HTML]{DAE8FC}\textbf{BCAM, TCAM, Associative} & N/A                                                         & Fixed                                                                     & User defined                                                              & One-to-One                                  & N/A                                          & N/A                                        & Low                                                                                                            \\ \hline
\cellcolor[HTML]{DAE8FC}\textbf{NS (Proposed)}        & Continuous                                                  & \begin{tabular}[c]{@{}c@{}}Changes Based on\\ Access Pattern\end{tabular} & \begin{tabular}[c]{@{}c@{}}Changes Based on\\ Access Pattern\end{tabular} & Many-to-Many                                & Many-to-Many                                 & Many-to-Many                               & High                                                                                                           \\ \hline
\end{tabular}
\end{table*}

In summary, we make the following contributions:
\begin{enumerate}
    \item We present a new paradigm of learning computer memory, called NS, that can track data access patterns to dynamically organize itself for providing high efficiency in terms of data storage and retrieval performance. We describe the learnable parameters and the learning process, which is incorporated into the store, retrieve and retention operations. 
    \item For quantitatively analyzing the capabilities of NS, we present a memory performance simulator with an array of tunable hyperparameters. 
    \item We present a formal process to select and customize NS for a target application. We also provide a comprehensive performance analysis of NS using two separate datasets representative of real-life applications and demonstrate its merit compared to traditional memory.
\end{enumerate}

 
The rest of the paper is organized as follows:
Section~\ref{sec:bckMotivation} discusses different state-of-the-art digital memory frameworks and provides motivations for the proposed intelligent digital memory design.
Section~\ref{Sec:BINGO} describes in details the proposed memory framework. Section~\ref{Sec:CaseStudy} quantitatively analyzes the effectiveness of the NS framework through multiple case-studies. Section~\ref{conclusion} concludes the main paper. Appendix~\ref{hyperparameter_appendix} provides additional details about the proposed hyperparameters. Appendix~\ref{opAlgo_appendix} provides detailed algorithms for different NS operations/procedures referenced in the main paper.
Appendix~\ref{expt_details_appendix} provides details about the experimental setup and hyperparameters used during the case studies. Appendix~\ref{dynamism_appendix} analyzes the dynamism of NS in greater depth. Appendix~\ref{addApp} discusses different applications which may benefit from using NS.

\section{Background and Motivation}
In this section, we shall first discuss the major difference between our propose memory framework (NS) and existing similar technologies. After that, we will provide motivations that led to the development of NS. 

\label{sec:bckMotivation}
\subsection{Computer Memory: A Brief Review}
\label{sec:relatedWorks}
Computer memory is one of the key components of a computer system \cite{hennessy2011computer_compArch}. Different types of memory have been proposed, implemented and improved over the decades. However, digital memories can still be can be broadly divided into two categories based on how data is stored and retrieved: (1) address operated and (2) content operated \cite{comSurveyBhunia}. In an address operated memory (for example a Random Access Memory or RAM \cite{hennessy2011computer_compArch, ram}), the access during read/write is done based on a memory address/location. During data retrieval/load, the memory system takes in an address as input and returns the data associated with it. Different variants of RAM such as SRAM (Static Random Access Memory) and DRAM (Dynamic Random Access Memory) are widely used \cite{hennessy2011computer_compArch}. In a content operated memory, on the contrary, memory access during read/write operations is performed based on a search pattern (i.e. content).


\begin{figure}[h!]
\centering
\includegraphics[width=\columnwidth]{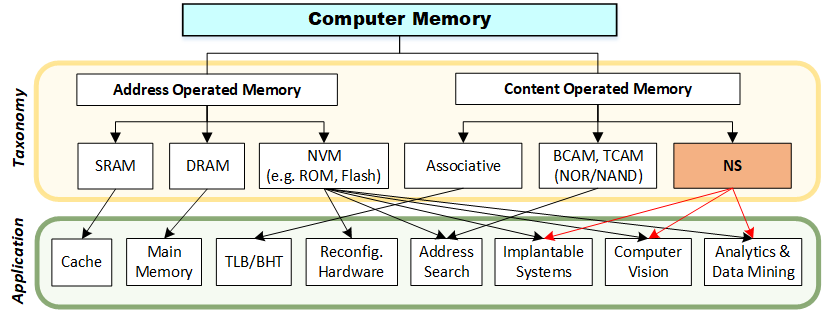}
\caption{Taxonomy of content operated memory used in computer systems. The proposed memory organization falls under the content-addressable memory category and is suitable for diverse application domains as shown.}
\label{fig:MemorySystemTypes}
\end{figure}
\begin{figure*}[h!]
\centering
\includegraphics[width=0.9\textwidth]{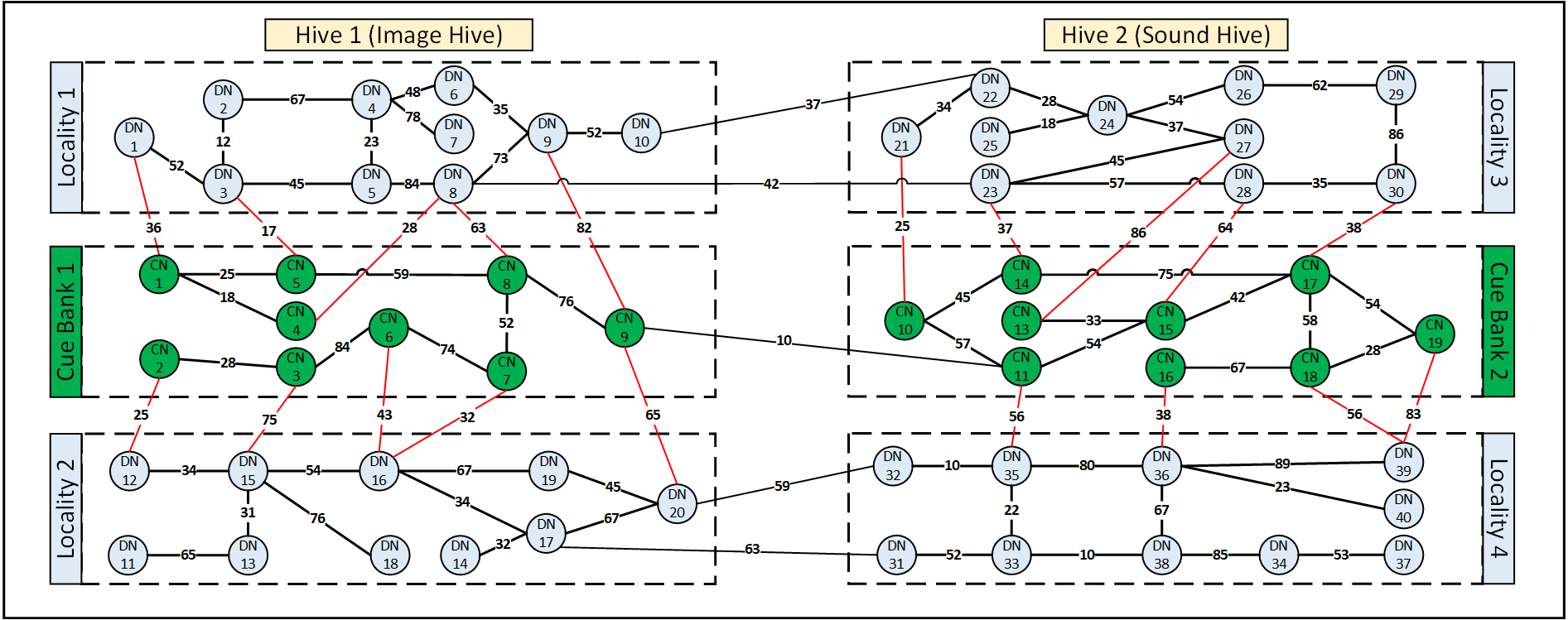}
\caption{Memory organization of NS with two memory hives. Data neurons (DN) store data and cue neurons (CN) store cues/tags. Each memory hive stores a specific data type (e.g., image, sound etc.). Localities are designed to store data consisting of specific features. We refer to this memory organization, the weighted graph of DN(s) and CN(s), as Neural Memory Network (NMN).}
\label{fig:system}
\end{figure*}

A COM (Content Operated Memory) \cite{com_survey, comSurveyBhunia} does not assign any specific data to a specific address during the store operation. During data retrieval/load, the user provides the memory system with a search pattern/tag and the COM searches the entire memory and returns the address in the memory system where the required data is stored. This renders the search process extremely slow if performed sequentially. To speed up this process of content-based searching, parallelization is employed which generally requires additional hardware. And adding more hardware makes the COM a rather expensive solution limiting its large-scale usability. A COM can be implemented in several ways as shown in Fig.~\ref{fig:MemorySystemTypes}, each with its own set of advantages and disadvantages. In an associative memory, the data are stored with a varying degree of restrictions. In a direct-mapped memory, each data can only be placed in one specific memory location. The restriction is less stringent in case of set-associative memory and in-case of fully associative memory, any data can reside anywhere in the memory. On the other hand, Neuromorphic associative memory behaves in a similar way as the standard associate memory at a high level but at a low-level, it exploits device properties to implement neuronal behaviour for increased efficiency \cite{pershin2011neuromorphic}. A CAM (Content-Addressable Memory) is similar to an associative memory in regards to its read and write behaviour however the implementation is different. In COM, there is a requirement for replacing old data units with new incoming data units in-case the memory runs out of space. The data unit/block to replace is determined based on a predefined replacement policy. 
CAM is the most popular variant of COM and is being used for decades in the computing domain but the high-level architecture of a CAM has not evolved much. Instead, researchers have mostly focused on how to best physically design a CAM to improve overall efficiency. SRAM bitcells are used as a backbone for any CAM cell \cite{com_survey, comSurveyBhunia}. Extra circuitry is introduced to perform the parallel comparison between the search pattern and the stored data. This is typically implemented using an XOR operation. The high degree of parallelism increases the circuit area and power overheads along with the cost. Cells implemented using NAND gates are more energy-efficient at a cost of speed. NOR gate based cells are faster but more energy-intensive.

Traditional CAMs are designed to be precise \cite{comSurveyBhunia}. No data degradation happens over time and in most cases, a perfect match is required with respect to the search pattern/tag to qualify for a successful retrieval. This feature is essential for certain applications such as destination MAC address lookup for finding the forwarding port in a network device. However, there are several applications in implantable, multimedia, Internet-of-Things (IoT) and data mining which can tolerant imprecise storage and retrieval. Ternary Content Addressable Memory (TCAM) is the only COM which allows partial match using a mask and are widely used in layer 3 network switches.

NS and CAM are both content operated memory frameworks. However, there are several differences between a traditional CAM and NS and some of the important ones are highlighted in Table~\ref{memComp}. For both Binary Content Addressable Memory (BCAM) and Ternary Content Addressable Memory (TCAM): (1) there are no learning components, (2) data resolution remains fixed unless directly manipulated by the user, (3) associations between search-pattern (tag/cue) and data remain static unless directly modified, (4) only a one-to-one mapping relation exists between search-pattern/cue and data units. Consequently, space and data fetch efficiency is generally low and we provide supporting results for this claim in Section~\ref{Sec:CaseStudy}.

Apart from standard computer memory organizations, researchers have also looked into different software level memory organizations for efficient data storage and retrieval. Instance retrieval frameworks are some such software wrappers on top of traditional memory systems that are used for feature-based data storage and retrieval tasks \cite{instRetSurvey}. These systems are mostly used for storing and retrieving images. During the training phase (code-book generation), visual words are identified/learned based on either SIFT features or CNN features of a set of image data. These visual words are, in most cases, cluster centroids of the feature distribution. Insertion of data in the system follows and is generally organized in a tree-like data structure. The location of each data in this data structure is determined based on the visual words (previously learned) that exist in the input image. During the retrieval phase, a search-image is provided and in an attempt to search for similar data in the framework, the tree is traversed based on the visual words in the search image. If a good match exists between the search image and a stored image, then that specific stored image is retrieved. The learning component in an instance retrieval frameworks is limited to the code-book generation phase which takes place during initialization. Furthermore, once a data unit is inserted in the framework, no more location and accessibility change is possible. No association exists between data units and granularity of data units do not change. On contrary, the overall dynamism and possibilities of a NS framework are much more. 

Another software level memory organization proposed by Niederee et. al. outlines the benefit of forgetfulness in a digital memory\cite{forgetfulMemory}. However, due to the lack of quantitative analysis and implementation details, it is unclear how effective this framework might be. 

\subsection{Motivation: Taking Inspiration from Human Memory}
\label{sec:motivation}

Computer and human memory are both designed to perform data storage, retention and retrieval. Although the functioning of human memory is far from being completely formalized and understood, it is clear that it is vastly different in the way data is handled. Several properties of the human brain have been identified which allows it to be far more superior than traditional computer memory in certain aspects. We believe that if some of these properties can be realized in a digital computer memory then many applications can benefit greatly. In the following subsections, we will look into some of the most important properties of the human brain and envision their potential digital counterparts. 

\subsubsection{Virtually Infinite Capacity}
The capacity of the human brain is difficult to estimate. John von Neumann, in his book \textit{``The computer and the brain"} \cite{von2012computer_computerAndTheBrain}, estimated that human brain has a capacity of $10^{20}$ bits with the assumptions: (1) All the inputs to the brain in its entire lifetime are stored forever, and (2) there are $10^{10}$ neurons in our brain. Researchers now even believe that our working memory (short-term memory) can be increased through ``\textit{plasticity}", provided certain circumstances. According to Lövd\'en et al., \textit{``... increase in working-memory capacity constitutes a manifestation of plasticity ..."} \cite{lovden2010theoretical_plasticity}. On top of that, due to intelligent pruning of unnecessary information, a human brain is able to retain only the key aspects of huge chunks of data for a long period of time. 

If a digital memory can be designed with this human brain feature, then the computer system, through intelligent dynamic memory re-organization (learning-guided plasticity) and via pruning features of unnecessary data (learned from statistical feedback), can attain a state of virtually infinite capacity. For example, in a wildlife image-based surveillance system which is geared towards detecting wolves, the irrelevant data (non-wolf frames) can be subject to compression/feature-loss to save space without hampering the effectiveness of the application.


\subsubsection{Imprecise/Imperfect Storage and Access}
The idea of pruning unnecessary data, as mentioned in the previous section, is possible because the human brain operates in an imprecise domain in contrary to most traditional digital memory frameworks. 
Human brain retrieval operation is imprecise in most situations\cite{atkinson1968human_memoryDecay} but intelligent feature extraction, analysis, and post-processing almost nullify the effect of this impreciseness. Also, certain tasks may not require precise memory storage and recall. For these tasks only some high-level feature extracted from the raw data is sufficient. 

Hence, supporting the imprecise memory paradigm in a digital memory is crucial for attaining the virtually infinite capacity and faster data access. For example, a wildlife image-based surveillance system can operate in the imprecise domain because some degree of compression/feature-reduction of images will not completely destroy the high-level features necessary for its detection tasks. This can lead to higher storage and transmission efficiency.

\subsubsection{Dynamic Organization}
We have mentioned that plasticity can lead to increased memory capacity but it also provides several other benefits in the human brain. According to Lindenberger et al., \textit{``Plasticity can be defined as the brain’s capacity to respond to experienced demands with structural changes that alter the behavioural repertoire."} \cite{lindenberger2017towards_plasticity}. Hence plasticity leads to better accessibility of important and task-relevant data in the human brain.
And the ease-of-access of a particular memory is adjusted with time-based on an individual's requirements. This idea is also similar to priming \cite{tulving1990priming_primingHelps} and it was observed that priming a human brain with certain memories helps in quicker retrieval.

If we can design a digital memory which can re-organize itself based on data access patterns and statistical feedback, then there will be great benefits in terms of reducing the overall memory access effort. For example, a wildlife image-based surveillance system which is geared towards detecting wolves will have to deal with retrieval requests mostly related to frames containing wolves. Dynamically adjusting the memory organization can enable faster access to data which are requested the most.

\subsubsection{Learning Guided Memory Framework}
Ultimately, the human brain can boast of so many desirable qualities mainly due to its ability to learn and adapt. 
It is safe to say, the storage policies of the human brain also vary from person to person and time to time \cite{atkinson1968human_memoryDecay}. Depending on the need and requirement, certain data are prioritized over others. The process of organizing the memories, feature reduction, storage and retrieval procedure changes over time based on statistical feedback. This makes each human brain unique and tuned to excel at a particular task at a particular time. 

Hence, the first step towards mimicking the properties of the human brain is to incorporate a learning component in the digital memory system. We envision that using this learning component, the digital memory will re-organize itself over time and alter the granularity of the data to become increasingly efficient (in terms of storage, retention and retrieval) at a particular task. For example, a wildlife image-based surveillance system which is geared towards detecting wolves will greatly benefit from a memory system which can learn to continuously re-organize itself to enable faster access to application-relevant data and continuously control the granularity of the stored data depending on the evolving usage scenario.

\section{Neural Storage Framework}
\label{Sec:BINGO}
To incorporate dynamism and embody the desirable qualities of a human brain in a digital memory, we have designed NS. It is an intelligent, self-organizing, virtually infinite content addressable memory framework capable of dynamically modulating data granularity. We propose a novel memory architecture, geared for learning, along with algorithms for implementing standard operations/tasks such as store, retrieve and retention. 

\subsection{Memory Organization}
The NS memory organization can be visualized as a network and we refer to it as ``Neural Memory Network" (NMN). The NMN, as shown in Fig.~\ref{fig:system} consists of multiple hives each of which is used to store data of a specific modality (type). For example, if an application requires to store image and audio data, then the NS framework will instantiate two separate memory hives for each data modality. This allows the search to be more directed based on the query data type. It is hypothesized that human memories are grouped up together to form small localities based on data similarity. We capture this idea by creating small memory localities within each hive that are designed to store similar data. The fundamental units of the NMN are (1) cue neurons and (2) data neurons. Each cue neuron stores a cue (data search pattern or tag) and each data neuron stores an actual data unit. Each data neuron is associated with a number denoting its `memory strength' which governs the data feature details or quality of the data inside it. A cue is essential a vector representing a certain concept and it can be of two types: (1) Coarse-grained cue and (2) Fine-grained Cue. Coarse-grained cues are used to navigate the NMN efficiently while searching (data retrieve operation) for a specific data and while navigating, the fine-grained cues are used to determine the data neuron(s) which is/are suitable for retrieval. A cue is a vector representing a particular concept but for the sake of simplicity, we shall use specific words when referring to certain cues. For example, in a wildlife surveillance system, cue neurons may contain vectors corresponding to a ``Wolf", ``Deer", etc but when talking about these cue-neurons we shall refer to them directly with the name of the concept they represent. The data neurons, for this example, will be image frames containing wolves, deer, jungle background, etc. Furthermore, if the system is designed to detect wolves, then the NS framework can be configured to have a memory locality for wolf-frames and one for non-wolf frames.


Each hive comes with its own Cue Bank which stores cue neurons arranged as a graph. The cue neuron and data neuron associations ($<$cue neuron, cue neuron$>$ and $<$cue neuron, data neuron$>$, $<$data neuron, data neuron$>$) change with time, based on the memory access pattern and framework hyperparameters. To facilitate multi-modal data search, connections between data neurons across memory hives are allowed. For example when searched with the cue ``Wolf" (the visual feature of a wolf), if the system is expected to fetch both images and sound data related to the concept of ``Wolf", then this above-mentioned flexibility will save search effort.

It is important to note that, the entire memory organization can be viewed as a single weighted graph where each node is either a data neuron or a cue neuron. The associations in the NMN are strengthened and weakened during store, retrieve and retention operations. With time, new associations are also formed and old associations may get deleted. The data neuron memory strengths are also modulated during memory operations to increase storage efficiency. The rigidity provided by hives, localities and cue-banks can be adjusted based on the application requirements.

\subsection{Parametric Space}
We propose several parameters for NS to help modulate how it functions. These parameters are of two types: (1) \textbf{Learnable Parameters} which changes throughout the system lifetime guided by reinforcement-learning and (2) \textbf{Hyperparameters} which are determined during system initialization and changed infrequently by the memory user/admin.

\subsubsection{Learnable Parameters}
We consider the following parameters as learnable parameters for NS:

\begin{enumerate}
    \item \textbf{Data Neuron and Cue Neuron Weighted Graph:} The weighted graph (NMN) directly impacts the data search efficiency (time and energy). The elements of the graph adjacency matrix are considered as learnable parameters. If there are $D$ number of data neuron and $C$ number of cue neuron at any given point of time, then the graph adjacency matrix will be of dimension $(D+C, D+C)$ .
    
    \item \textbf{Memory Strength Array:} The memory strengths of all the data neurons are also considered as learnable parameters. They jointly dictate the space-utilization, transmission efficiency and retrieved data quality.
    
\end{enumerate}

These parameters constantly evolve based on the system usage via a reinforcement learning process. We will do a deeper dive into the learning process in Section~\ref{Sec:learning}.

\subsubsection{Hyperparameters}
\label{hyperparameter}
We have also defined a set of hyperparameters which influences the NS memory organization and the learning guided operations.
These hyperparameters can be set/changed by the user during the setup or during the operational stage. The first hyperparameter is the \textbf{number of memory hives} and we propose the following hyperparameters for each \textbf{hive}:
\begin{enumerate}
    \item \textbf{Number of localities}: Each locality is used to store a specific nature of data. It is an unsigned integer value.
    \item \textbf{Memory decay rate of each locality}: Controls the rate at which data neuron memory strength and features are lost due to inactivity. 
    \item \textbf{Association decay rate of each locality}: Controls the rate at which NMN associations losses strength due to inactivity.
    \item \textbf{Mapping between data features and localities}: This mapping dictates the segregation of application relevant data and their assignment to a locality with a low decay rate.
    \item \textbf{Data features and cue extraction AI (Artificial Intelligence) models}: These models are used to obtain more insights about the data. They should be selected based on the application and data-type being processed. 
    
    \item \textbf{Data neuron matching metric}: Used during retrieve operation for finding a good match and during store operation for data neuron merging. For example, this metric can be something like \textbf{cosine similarity}.

    \item \textbf{Neural elasticity parameters}: Determines the aggressiveness with which unused data neurons are compressed in-case of space shortage.
    
    \item \textbf{Association weight adjustment parameter}: Used as a step size for increasing/decreasing association weights inside the NMN. A higher value will increase the dynamism but lower the stability.
    
    \item \textbf{Minimum association weight ($\varepsilon$)}: It is an unsigned integer which limits the decay of association weight beyond a certain point.
  
    \item \textbf{Degree of impreciseness ($\varphi$)}: Limits the amount of data feature which is allowed to be lost due to memory strength decay and inactivity. It is a floating-point number in the range [ 0 - 100 ]. 0 implies data can get completely removed if needs arise.
    
    \item \textbf{Frequency of retention procedure ($N$)}: NS has a retention procedure which brings in the effect of ageing. This hyperparameter is a positive integer denoting the number of normal operations to be performed before the retention procedure is called once. A lower value will increase dynamism at a cost of overall operation effort (energy and time consumption).
    
    \item \textbf{Compression techniques}: For each memory hive we must specify the algorithm to be used for compressing the data when required. For example, we can use JPEG \cite{ref:jpeg} compression in an image hive.
    
\end{enumerate}

More details can be found in Appendix~\ref{hyperparameter_appendix}. 

\begin{figure}[h!]
\centering
\includegraphics[width=\columnwidth]{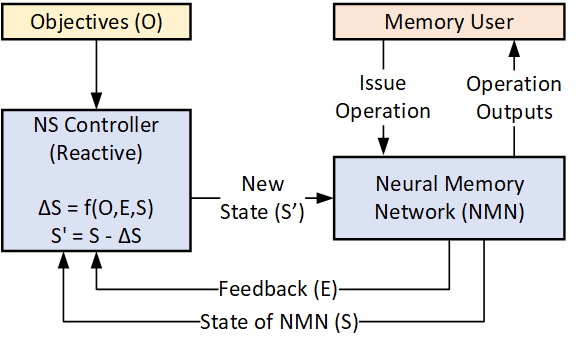}
\caption{The proposed reinforcement learning architecture used to incorporate learning in the NS framework. Each memory operation generates a feedback (E), which is used to modify and optimize the NMN by changing its current state $S$ to the newly computed state $S$'.}
\label{fig:feedbackloop}
\end{figure}

    


\subsection{Learning Process}
\label{Sec:learning}
The \textbf{learnable parameters}, governing the NMN of NS are updated based on feedback from memory operations. The goal/objective of learning is to:
\begin{enumerate}
    \item Reduce space requirement while maintaining data retrieval quality and application performance. This should be achieved by learning the granularity at which each data neuron should be stored. Less important data should be compressed and subject to feature-loss for saving space while more important data should be kept at a good quality. Hence this learning should be driven based on the access-pattern.
    \item Increase memory search speed by learning the right NMN organization given current circumstances and access-pattern bias.
\end{enumerate}

In Fig.~\ref{fig:feedbackloop}, we propose an external stimulus guided reactional reinforcement learning (RL) architecture for incorporating intelligence in NS. The NS framework consists of two main components: (1) The Neural Memory Network (NMN) and (2) The NS Controller which manages the NMN.

The initial state ($S_0$) of the NMN consists of no data-neuron (DN) and no cue-neuron (CN). During an operation when a new cue is identified (not present in the cue bank), a new cue-neuron (CN) is generated for that cue. Similarly, when an incoming data cannot be merged with an existing data-neuron (DN), a new DN is created. Each new DN is initialized with a memory strength of 100\% (this parameter dictates the data granularity/details for the DN). When a new DN or CN is created, the new neuron is connected with all other exiting neurons (DNs and CNs) with an association weight of $\varepsilon$ (a hyperparameter selected by the system admin/user). So in any state, all DNs and CNs form a fully connected weighted graph. 

At the end of each operation, a feedback (E) is generated and sent to the NS Controller module along with the snapshot of the current state of the NMN (S). $S$ (essentially the \textbf{Learnable Parameters}) has two components:
\begin{enumerate}
    \item \textbf{$S \rightarrow A$}: The adjacency matrix for the entire NMN.
    \item \textbf{$S \rightarrow M$}: The list of memory strengths of each data neuron.
\end{enumerate}

For an NMN with $n$ total neurons (DNs and CNs) and $m$ number of DNs:
$$ S \rightarrow A = 
\begin{bmatrix} 
a_{11} & a_{12} & a_{13} & ... & a_{1n} \\
a_{21} & a_{22} & a_{23} & ... & a_{2n} \\
... & ... & ... & ... & ... \\
a_{n1} & a_{n2} & a_{n3} & ... & a_{nn} \\
\end{bmatrix}
\quad
$$

$$ S \rightarrow M = 
\begin{bmatrix} 
s_{1} & s_{2} & s_{3} & s_{4} & s_{5} & ... & s_{m} \\
\end{bmatrix}
\quad
$$

$S$ and $E$, along with the learning goals/objectives (O) drives the reaction function $f(O,E,S)$. The outputs of this function are:
\begin{enumerate}
    \item An association weight adjustment matrix ($\Delta A$) of dimension $(n, n)$.
    \item A memory strength adjustment vector ($\Delta M$) of dimension $(1, m)$.
\end{enumerate}

These 2 components constitute $\Delta S = \{\Delta A, \Delta M\}$.
$$ \Delta A = 
\begin{bmatrix} 
\delta a_{11} & \delta a_{12} & \delta a_{13} & ... & \delta a_{1n} \\
\delta a_{21} & \delta a_{22} & \delta a_{23} & ... & \delta a_{2n} \\
... & ... & ... & ... & ... \\
\delta a_{n1} & \delta a_{n2} & \delta a_{n3} & ... & \delta a_{nn} \\
\end{bmatrix}
\quad
$$

$$ \Delta M = 
\begin{bmatrix} 
\delta s_{1} & \delta s_{2} & \delta s_{3} & \delta s_{4} & \delta s_{5} & ... & \delta s_{m} \\
\end{bmatrix}
\quad
$$

We compute $S'$ as follows: 

$$ S' \rightarrow A = 
\begin{bmatrix} 
max(\varepsilon, a_{11}-\delta a_{11}) & ... & max(\varepsilon,a_{1n}-\delta a_{1n}) \\
max(\varepsilon,a_{21}-\delta a_{21})  & ... & max(\varepsilon,a_{2n}-\delta a_{21}) \\
... &  ... & ...  \\
max(\varepsilon,a_{n1}-\delta a_{n1})  & ... & max(\varepsilon,a_{nn}-\delta a_{n1}) \\
\end{bmatrix}
\quad
$$

$$ S' \rightarrow M = 
\begin{bmatrix} 
min(100, max(\varphi,s_{1}-\delta s_{1})) &\\ min(100, max(\varphi,s_{2}-\delta s_{2})) &\\ ... &\\ min(100,max(\varphi,s_{m}-\delta s_{m})) \\
\end{bmatrix}
\quad
$$

Where, $\varphi$ (degree of impreciseness, a hyperparameter selected by the system admin/user) is the minimum memory strength a data-neuron can have.

\begin{center}
\begin{figure*}[!h]
\centering
\begin{subfigure}[b]{\textwidth}
\centering
\includegraphics[width=\textwidth]{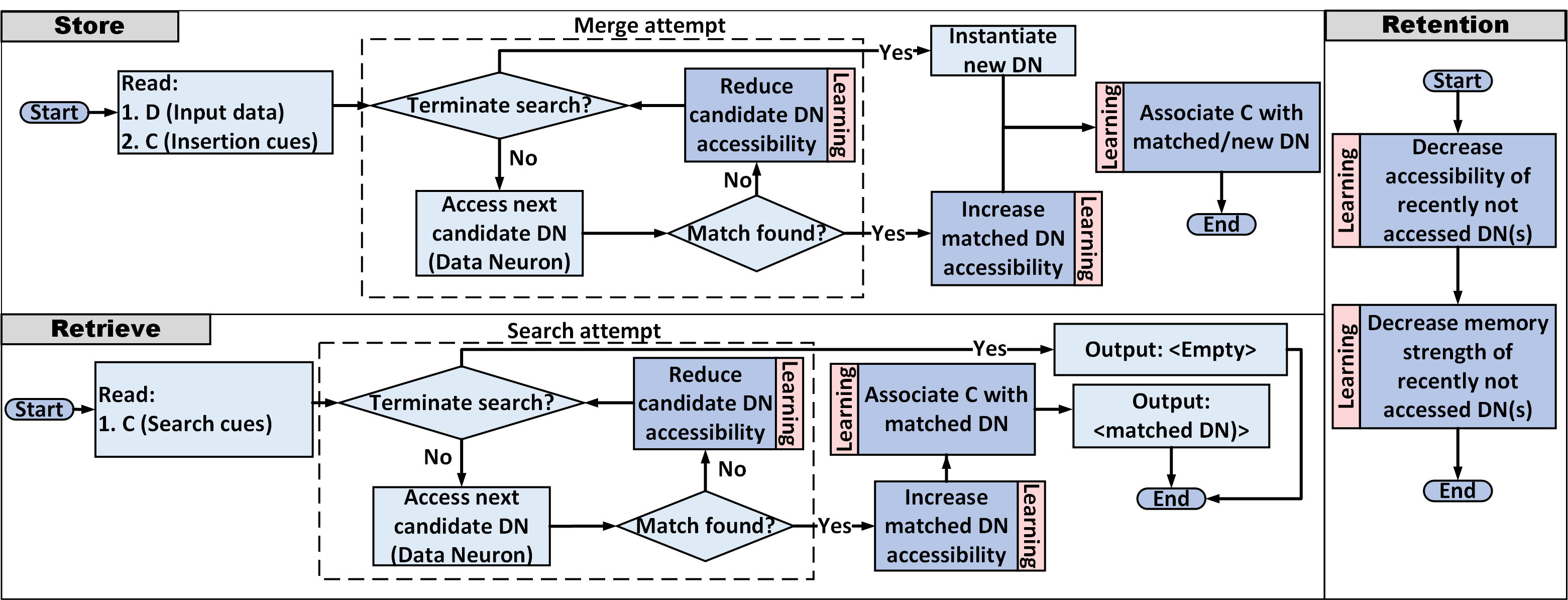}
\caption{}
\end{subfigure}
\begin{subfigure}[b]{\textwidth}
\centering
\includegraphics[width=\textwidth]{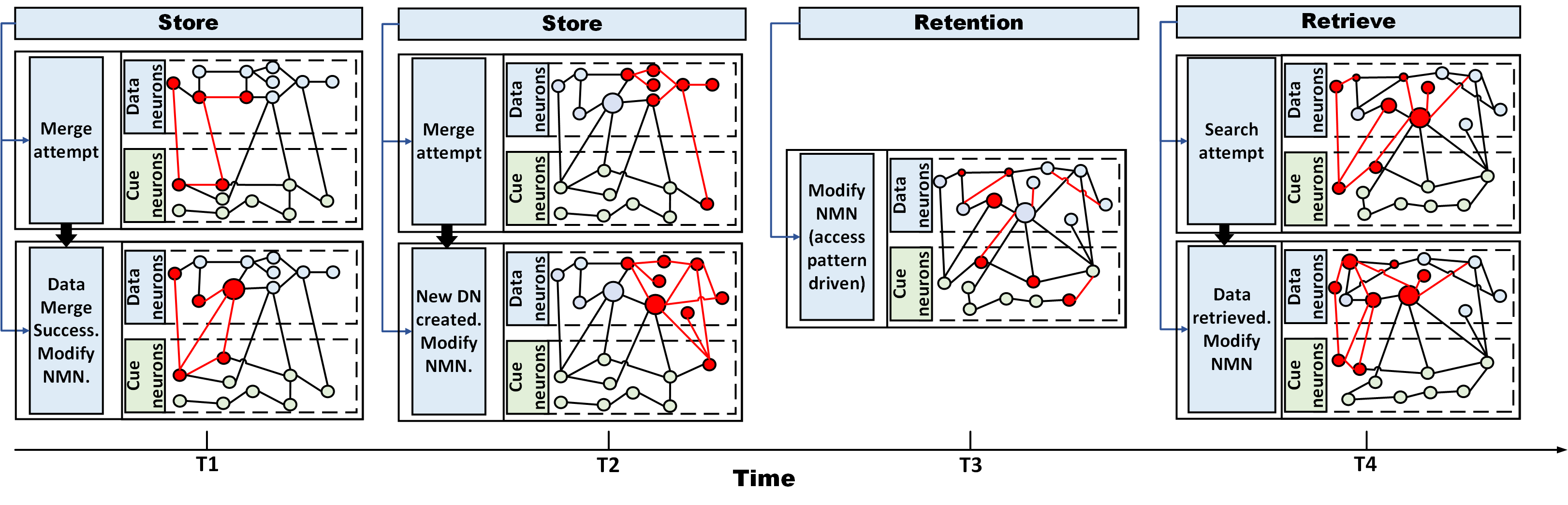}
\caption{}
\end{subfigure}
\begin{subfigure}[b]{\textwidth}
\centering
\includegraphics[width=\textwidth]{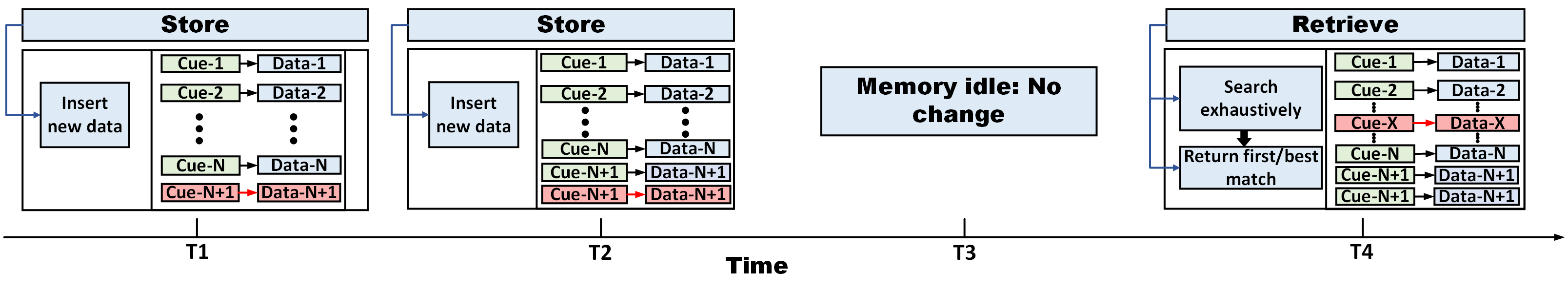}
\caption{}
\end{subfigure}

\caption{(a) Flowcharts illustrating the major steps of store, retrieve and retention operations in NS. (b) Visualization of the NS memory structure during a sequence of operations. (c) Visualization of a traditional CAM for an identical sequence of operations.}
\label{fig:op_HighLevel}
\end{figure*}
\end{center}

The memory state is updated with the newly computed one ($S'$). The function $f(O,E,S)$ for computing $\Delta M$ and $\Delta A$ can be realized in many different ways depending on the implementation. The updates made to the matrices for a given state, $S$ can be made local in nature to reduce unnecessary computations and updates. The periodicity of the state update can also be controlled. For the current implementation of NS used for performing the case-studies, the reaction function is jointly implemented using Algo.~\ref{alg:store}, Algo.~\ref{alg:retrieve} and Algo.~\ref{alg:reaction}. The algorithms are discussed in Appendix~\ref{opAlgo_appendix} and the high level concept is provided in Fig.\ref{fig:op_HighLevel} (a).

\subsection{Memory Operations}
\label{Sec:op}
We have designed three fundamental NS memory operations (Store, Retrieve, and Retention) which are analogous to similar operations that exist in any traditional CAM. The learning process is directly embedded as parts of these operations. The flowcharts of these operations are shown in Fig.~\ref{fig:op_HighLevel} (a) and the implementation level details can be found in Appendix~\ref{opAlgo_appendix}.

\subsubsection{Store}
The store operation (as shown in Fig.~\ref{fig:op_HighLevel} (a)) starts by reading the input data (D) and insertion cues (C). Before storing the input data using a new data neuron, the NS framework attempts to merge it with an existing data neuron with similar content. This sub-operation (merge attempt) is designed to eliminate storing similar data multiple times. During the merge attempt, a set of candidate data neurons (selected based on accessibility with respect to the insertion cues, C) are examined for a good match and the data neurons that do not match are penalized by being made less accessible in the NMN. If a data neuron having a good match with the input data (D) is found, then that matching data neuron is assigned a higher memory strength and made more accessible in the NMN. After the merge attempt, if a good match is not found, a new data neuron is instantiated for the input data (D). If a new cue (not present in the cue bank) is found among C, then a new cue-neuron is instantiated for it. Depending on the merge attempt success/failure, the new data neuron or the matching data neuron respectively is associated (if already associated, then strengthened) with the insertion cues (C). 

The learning aspect of this operation is guided by the input data (D) and cues (C) provided. The candidate data neurons for merging are selected using a graph traversal starting from the insertion cues (C). The graph traversal is guided by the NMN structure, hence wrong candidate data neuron selections are penalized by making those candidate data neurons less accessible (by NMN modification). On the other hand, selecting a candidate data neuron with a good match, with respect to the input data, is rewarded by making that candidate data neuron more accessible (by NMN modification). Association of the insertion cues (C) with the matching data neuron or the new data neuron can also be considered as a learning process.

\subsubsection{Retrieve}
The retrieve operation (as shown in Fig.~\ref{fig:op_HighLevel} (a)) starts by reading the search cues (C). The search cues consist of a set of coarse-grained cues ($C_1$) and, optionally, a set of fine-grained cues ($C_2$). Based on $C_1$, a set of candidate data neurons are selected and are checked for an acceptable match with respect to the fine-grained cues. The candidate data neurons that do not match with any fine-grained cue in $C_2$ are made less accessible in the NMN and if a candidate data neuron matches with any fine-grained cue in $C_2$, then it is made more accessible in the NMN. At the end of the search attempt, if a matching data neuron is located, then it is provided as output and it also gets associated with all the search cues (C) inside the NMN. In absence of $C_2$, the retrieve operation returns the first accessed candidate data neuron during the search phase.

Similar to the store operation, the learning in this operation is also driven by the candidate data neuron selection, which is primarily based on the NMN organization/structure. A wrong candidate selection is penalized and a good candidate selection is rewarded by making necessary NMN state modifications. Association of search cues (C) with the matching data neuron is also a part of the learning process.

\subsubsection{Retention}
In a traditional CAM, data retention involves maintaining the memory in a fixed state. NS, on the other hand, allows the NMN to change and restructure itself to show the effect of ageing as shown in Fig.~\ref{fig:op_HighLevel} (a). All the data neurons not accessed in the last N-operations (N is a hyperparameter selected by the system admin/user), are weakened. Weakening a data neuron leads to data feature loss. This sub-operation is a form of reinforcement learning which considers the access pattern and determines which data neurons to shrink for saving space. The next sub-operation is also learning-driven, where the accessibility of unused data neurons are reduced based on the access pattern. 

\subsection{Dynamic Behaviour of NS}
In comparison to traditional CAM, NS is dynamic in several aspects. The NMN of NS changes after every operation and the effect of ageing is captured using the retention procedure. In Fig.~\ref{fig:op_HighLevel} (b), we illustrate the dynamic nature of NS by displaying how the NMN changes during a sequence of operations. Accessibility of different data neurons are changed and memory strength of data neurons increase or decrease based on the feedback-driven reinforcement learning scheme. In contrast, as seen in Fig.~\ref{fig:op_HighLevel} (c), the traditional CAM does not show any sign of intelligence or dynamism to facilitate data storage/retrieval. In Appendix~\ref{dynamism_appendix}, a more detailed simulation-accurate depiction and description of NS's dynamism are provided.

\begin{figure*}[h!]
\centering
\includegraphics[width=0.9\textwidth]{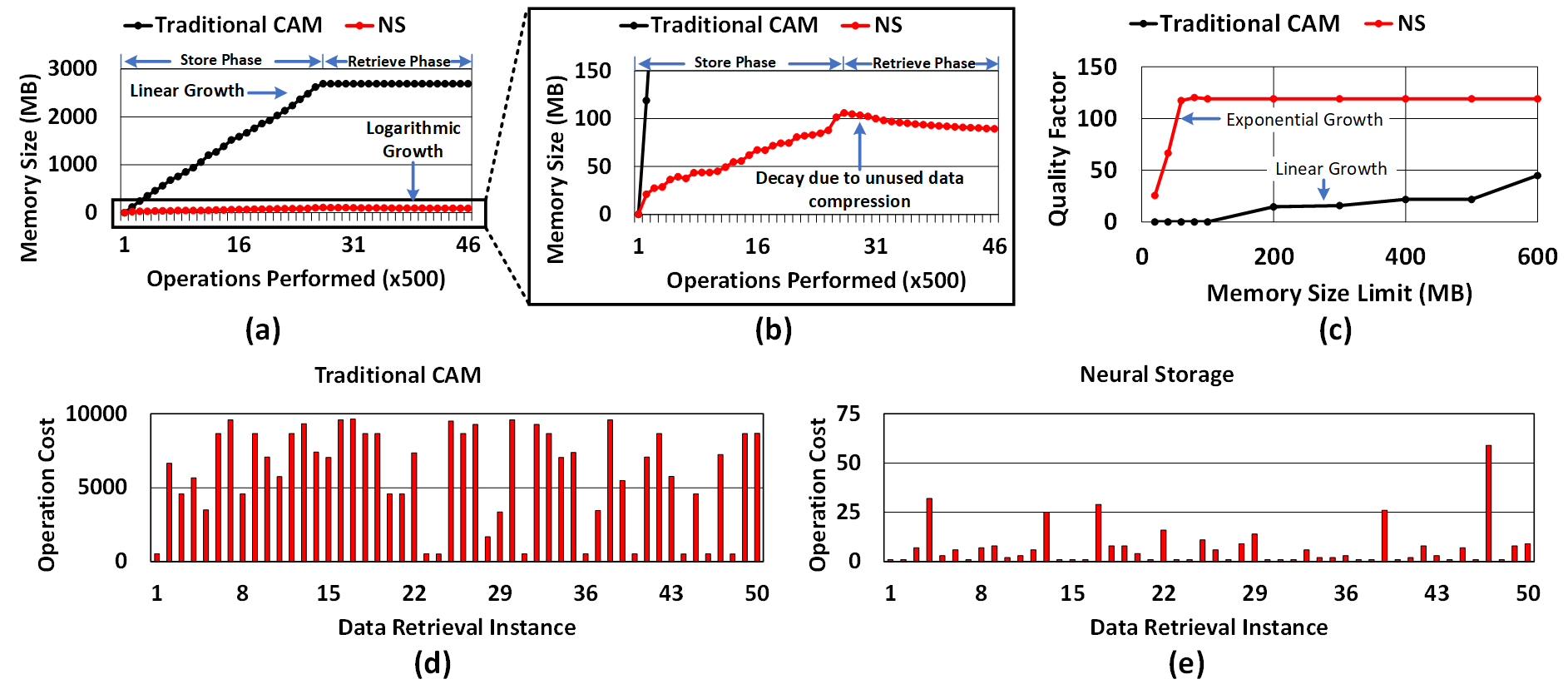}
\caption{Results for the wildlife surveillance application which prioritizes deer images (scenario 1). (a) This plot shows the memory growth for traditional CAM and NS. The NS memory growth is logarithmic in nature, while the Traditional CAM space utilization grows linearly. (b) The zoomed-in portion of the memory growth plot shows the space utilization fluctuation due to access-pattern driven learning guided data compression in NS framework. (c) This plot shows the memory quality factor for NS and Traditional CAM at different memory size limits. It is clear that NS can operate more efficiently in space-constrained scenarios. (d,e) Shows the retrieval operation costs for the NS and the traditional CAM. NS appears to be far superior due to dynamism.}
\label{fig:DeerRes}
\end{figure*}
\subsection{NS Simulator}
\label{bingoSimulator}
In order to quantitatively analyze the effectiveness of NS in a computer system, we have designed and implemented a NS simulator. It has the following features:

\begin{itemize}
    \item It can simulate all memory operations and provide relative benefits with respect to traditional CAM in terms of operation cost.
    \item The framework is configurable an array of hyperparameters introduced in Section~\ref{hyperparameter}.
    \item The NS simulator can be mapped to any application which is designed for using a CAM or a similar framework.
    \item The simulator implements the learning paradigm as shown in Fig.~\ref{fig:op_HighLevel} and more details can be found in Appendix~\ref{opAlgo_appendix}.
    \item The NS simulator is scalable and can simulate a memory of arbitrarily large size.
    \item To ensure correctness, the simulator software is validated through manual verification of multiple random case studies with a large number of random operations.
\end{itemize}

We define operation cost as the amount of effort it takes to perform a particular operation. It is clear that the iterative search-section of the NS operations (Fig.~\ref{fig:op_HighLevel}), dominates over the remaining sub-operational steps in terms of effort. Hence, we approximately consider the operation cost for NS to be the number of times the search-section is executed for both store and retrieve operations. For the traditional CAM, we consider the operation cost to be the number of data entries searched/looked-up. For both traditional CAM and NS, we do not consider any parallelism while searching to ensure fairness. Also, the cost of writing the data to the memory for both NS and traditional CAM is not considered as a part of the operation cost. For NS, the effort of writing the data to the memory is less than or equal (in the worst case, due to data merging) to that of the traditional CAM.

\begin{figure*}[h!]
\centering
\includegraphics[width=0.9\textwidth]{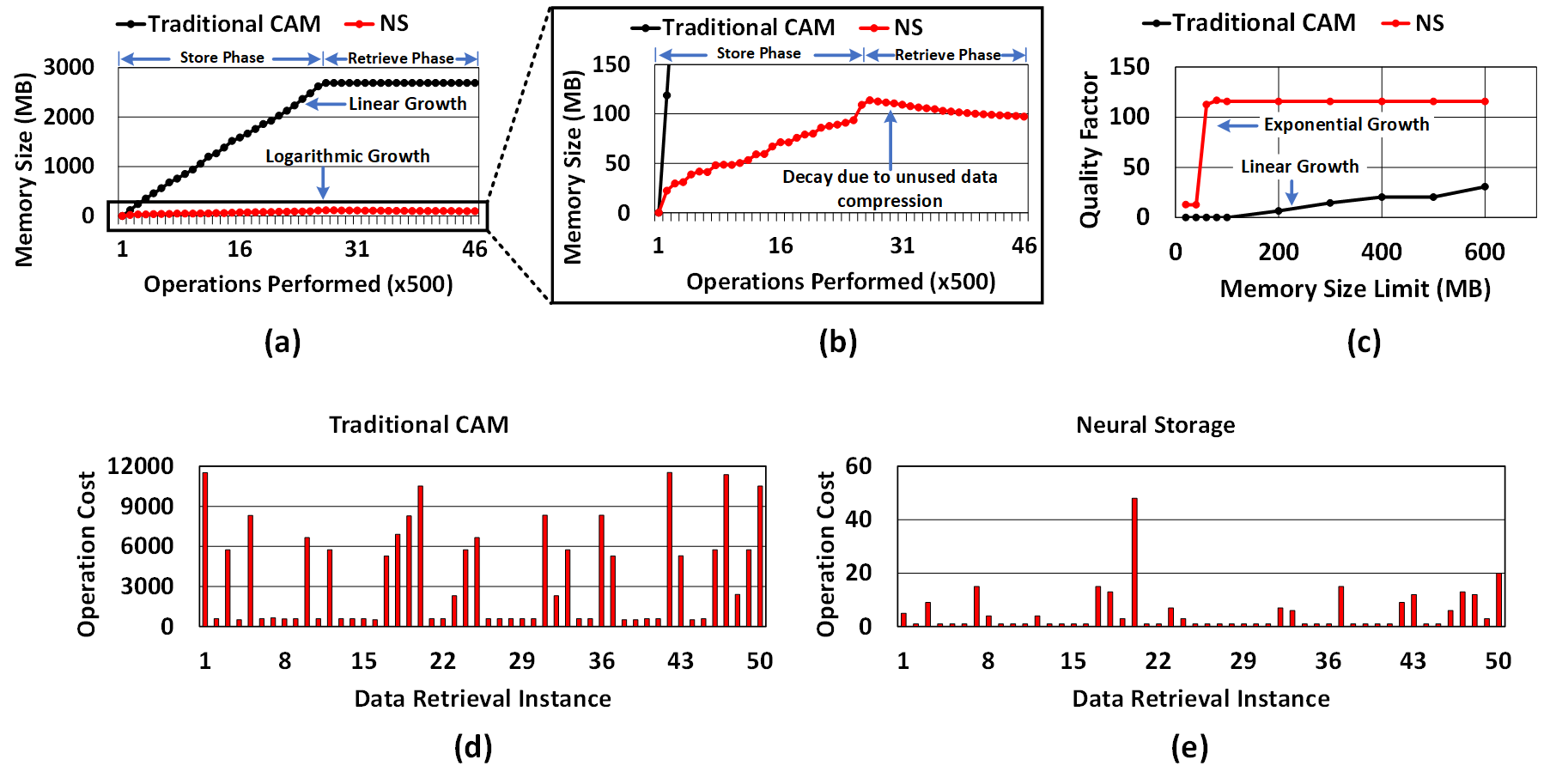}
\caption{Results for wildlife surveillance application which prioritizes fox/wolf images (scenario 2). We show similar plots and highlight similar results as Fig.~\ref{fig:DeerRes}. In summary, NS appears to be more efficient than traditional CAM.}
\label{fig:FoxRes}
\end{figure*}
\subsection{Desirable Application Characteristics}
\label{bingoTrait}
Any application using a CAM or a similar framework can be theoretically replaced with NS. However, certain applications will benefit more than others. The two main traits of an application which will enhance the effectiveness of NS are described below.


\subsubsection{Imprecise Store \& Retrieval}
Although NS can be configured to operate at 100\% data precision, it is recommended to use the framework in the imprecise mode for storage and search efficiency. Assume $D =$ Set of data neurons in the Memory at a given instance. For a given data $D_i \in M$, if $D_i$ is compressed (in lossy manner) to ${D_i}'$ and $(size(D_i) = size({D_i}') - \epsilon_1)$, then in order for the application to operate in the imprecise domain, it must be that $(Quality ({D_i}') = Quality ({D_i}') - \epsilon_2)$. Where size(X) is the size of the data neuron X and Quality(X) is the quality of the data in the data neuron X, in light of the specific application. $\epsilon_1$ and $\epsilon_2$ are small quantities. For example, in a wildlife surveillance system, if an image containing a wolf is compressed slightly, it will still look like an image with the same wolf. 

\subsubsection{Notion of Object(s)-of-Interest}
NS works best if there exists a set of localities within each hive which are designated to store data containing specific objects. Each locality can be configured to have different memory strength decay rate based on the importance of the data which are designated to be stored in the respective locality.
Note that the definition of an object in this context implies specific features of the data. For example, in case of an image data, the object can be literal objects in the image but for a sound data, the objects can be thought of as a specific sound segment with certain attributes. 
Assume that D is a new incoming data which must be stored in the Memory and $OL = $ objects in data D. Then there may be situations where, $\exists O_1, O_2 \in OL \mid  Imp(O\_1) > Imp(O\_2) $. Where $Imp(O\_i)$ denotes the importance of the object $O\_i$ for the specific application. For example, in wildlife surveillance designed to detect wolves, frames containing at least one wolf should be considered with higher importance.

\section{Case Study}

\label{Sec:CaseStudy}
To evaluate the effectiveness of NS, we choose two image datasets from two representative applications: (1) wildlife surveillance system and (2) a UAV-based security system. For both applications, traditional CAM can be used for efficient data storage and retrieval. We performed a comparative study between NS and traditional CAM for these datasets in terms of several key memory access parameters. To model NS behaviour for the target dataset, we use the simulator as described in Section \ref{bingoSimulator}, while traditional CAM behaviour is modelled based on standard CAM organization \cite{CAMSurvey_Imp}. Next, we present a quantitative analysis on the benefits of using NS over traditional CAM. The hyperparameters and configuration details of the NS simulator used in the case studies are described in Appendix~\ref{expt_details_appendix}.

\subsection{Wildlife Surveillance}
Image sensors are widely deployed in the wilderness for rare species tracking and poacher detection. The wilderness can be vast and IoT devices operating in these regions often deal with low storage space, limited transmission bandwidth and power/energy shortage. This demands efficient data storage, transmission and power management. Interestingly, this specific application is resistant to imprecise data storage and retrieval because compression does not easily destroy high-level data features in images. Also, in the context of this application, certain objects such as a rare animal of a specific species are considered more important than an image with only trees or an unimportant animal. Hence this application has the desirable characteristics suitable for using NS and will certainly benefit from NS's learning guided preciseness modulation and plasticity schemes. Informed reduction of unnecessary data features will also lead to less transmission bandwidth requirements. Memory power utilization is proportional to the time required to carry out a store, load and other background tasks. And NS, due to its efficient learning-driven dynamic memory organization, can help reduce memory operation time and consequently can reduce overall operation effort. Furthermore, transmitting the optimal amount of data (instead of the full data) will lead to lesser energy consumption as transmission power requirement is often much higher than computation \cite{sadler2006data}. 

\begin{figure*}[h!]
\centering
\includegraphics[width=0.9\textwidth]{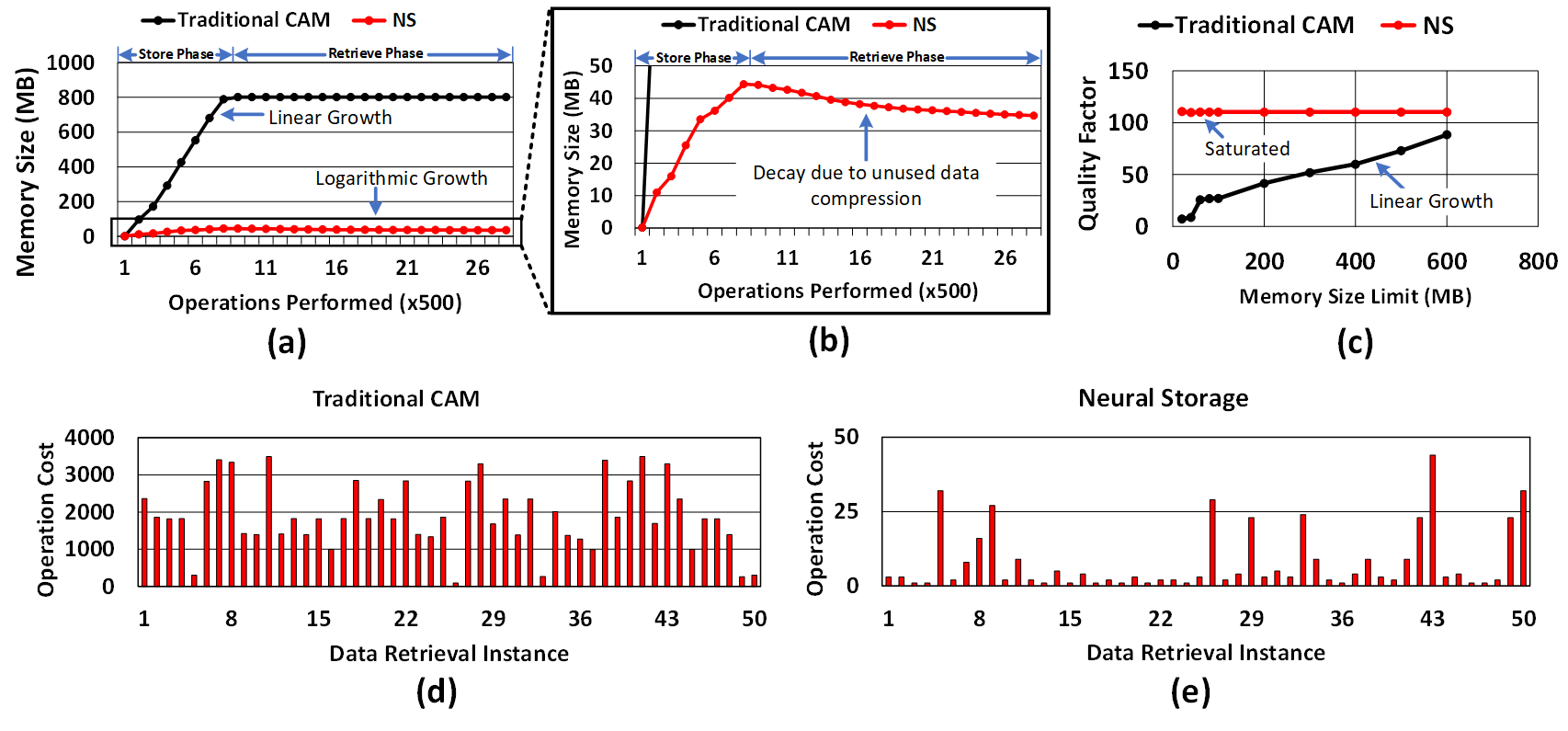}
\caption{Results for UAV-based car surveillance application which prioritizes car images. We show similar plots and highlight similar observations as Fig.~\ref{fig:DeerRes}. In summary, it appears that NS is more efficient than traditional CAM in terms of both storage and retrieval efficiency.}
\label{fig:CarRes}
\end{figure*}
\subsubsection{Dataset Description}
To emulate a wildlife surveillance application, we construct an image dataset from a wildlife camera footage containing 40 different animal sightings. The details, of this dataset, are located at the publicly available dataset repository that we have released \cite{ref:BINGO_DS}. We construct two different scenarios for carrying out experiments on this dataset:

\begin{itemize}
    \item \textbf{Scenario 1:} The system user wishes to prioritize deer images and perform frequent deer image retrieval tasks.
    \item \textbf{Scenario 2:} The system user wishes to prioritize fox/wolf images and perform frequent fox/wolf image retrieval tasks.
\end{itemize}

\begin{table*}[]
\centering
\caption{Simulation results for different case-studies showing the relative effectiveness of NS with respect to traditional CAM. Judging by the PSNR and SSIM of the retrieved images, it is clear that NS provides similar performance in terms of the quality of retrieved data with respect to traditional(Trad.) CAM. This leads us to believe that, the application performance will not be hampered from using NS. At the same time, NS is utilizing much less space and average operation cost is far lower in comparison to traditional CAM.}
\label{BINGO_collectedResTable}
\renewcommand{\arraystretch}{1.3}
\scriptsize\addtolength{\tabcolsep}{4pt}
\begin{tabular}{|c|c|c|c|c|c|c|}
\hline
\rowcolor[HTML]{EFEFEF} 
\multicolumn{7}{|c|}{\cellcolor[HTML]{EFEFEF}\textbf{Wildlife Surveillance: Emphasis on Deer (Scenario-1)}}                                      \\ \hline
\rowcolor[HTML]{ECF4FF} 
Mem. Type & PSNR & SSIM & Avg. Retrieval Op. Cost & \cellcolor[HTML]{ECF4FF}Avg. Store Op. Cost & \cellcolor[HTML]{ECF4FF}Avg. Op. Cost & Final Memory Size (MB) \\ \hline
\cellcolor[HTML]{FFFFC7}Trad. CAM & 37.65 & 0.79 & 5573.93                 & 0                   & 2786.96       & 2688.79          \\ \hline
\cellcolor[HTML]{FFFFC7}NS     & 38.39 & 0.82 & 5.51                    & 28.23               & 16.87         & 89.46            \\ \hline
\rowcolor[HTML]{EFEFEF} 
\multicolumn{7}{|c|}{\cellcolor[HTML]{EFEFEF}\textbf{Wildlife Surveillance: Emphasis on Fox (Scenario-2)}}                                       \\ \hline
\rowcolor[HTML]{ECF4FF} 
Mem. Type                         & PSNR  & SSIM & Avg. Retrieval Op. Cost & Avg. Store Op. Cost & Avg. Op. Cost & Final Memory Size (MB) \\ \hline
\cellcolor[HTML]{FFFFC7}Trad. CAM & 37.11 & 0.75 & 3306.83                 & 0                   & 1653.41       & 2688.79          \\ \hline
\cellcolor[HTML]{FFFFC7}NS     & 38.52 & 0.79 & 4.39                    & 22.96               & 13.67         & 97.40            \\ \hline
\rowcolor[HTML]{EFEFEF} 
\multicolumn{7}{|c|}{\cellcolor[HTML]{EFEFEF}\textbf{UAV-based Surveillance for Safety: Emphasis on Car}}                           \\ \hline
\rowcolor[HTML]{ECF4FF} 
Mem. Type                         & PSNR  & SSIM & Avg. Retrieval Op. Cost & Avg. Store Op. Cost & Avg. Op. Cost & Final Memory Size (MB) \\ \hline
\cellcolor[HTML]{FFFFC7}Trad. CAM & 30.57 & 0.71 & 1870.41                 & 0                   & 935.20        & 801.14           \\ \hline
\cellcolor[HTML]{FFFFC7}NS     & 32.42 & 0.78 & 6.03                    & 22.67               & 14.355        & 34.67            \\ \hline
\end{tabular}
\end{table*}
\subsubsection{Effectiveness of NS in Comparison to Traditional CAM}
Both the NS framework and the traditional CAM are first presented with all the images in the dataset sequentially and then a randomly pre-generated access pattern (based on the scenario) is used to fetch 10,000 images (non-unique) sampled from the dataset. For scenario-1, as can be seen in Fig.~\ref{fig:DeerRes} (a), NS has a clear advantage over traditional CAM in terms of total space utilization. We also observe in the zoomed-in graph, Fig.~\ref{fig:DeerRes} (b), the NS total space utilization fluctuates and slowly decreases after the end of the store phase. This is due to access pattern guided optimal data granularity learning resulting in compression/feature-loss of less important data. In Fig.~\ref{fig:DeerRes} (d) and Fig.~\ref{fig:DeerRes} (e), we plot the operation cost (as described in Section~\ref{bingoSimulator}) during the first 50 retrieve operations for traditional CAM and NS respectively. We observe that the operation cost for the NS in comparison to traditional CAM is significantly lower. 

In Table~\ref{BINGO_collectedResTable}, we present the numerical details of all the experiments. The average operation cost (store and retrieve combined) for NS is about 165 times less than that of traditional CAM. It is worth noting that the PSNR (Peak signal-to-noise ratio) and SSIM (Structural similarity) of the fetched images during retrieve operations for NS have similar values as that of traditional CAM. This ensures that using the NS framework for this application will not affect the effectiveness of the application. We next perform the same experiments with locality-0 tuned to store only fox/wolf frames (scenario-2). In Fig.~\ref{fig:FoxRes} and Table~\ref{BINGO_collectedResTable}, we observe similar trends.




\subsubsection{Effectiveness of NS in Constrained Settings}
Most of the image sensors used in a wildlife surveillance system are deployed in remote locations and must make efficient use of bandwidth and storage space without sacrificing system performance. NS is designed to excel in this scenario and to verify this, we limit the total memory size (X-axis) and plot the memory quality factor (Y-axis) in Fig.~\ref{fig:DeerRes} (c) (scenario-1) and Fig.~\ref{fig:FoxRes} (c) (scenario-2). The memory quality factor is defined in Eqn.~\ref{eqn1}. In both the scenarios we observe that NS is capable of functioning at a much lower memory capacity in comparison to Traditional CAM. Lower space utilization also translates to less transmission bandwidth consumption, in-case the system has to upload the stored data to the cloud or other IoT devices. Also, note that the quality factor of the NS framework increases exponentially with the increase in memory size limit whereas the quality factor of the traditional CAM increases at a much slower pace. 
\begin{equation}
\label{eqn1}
Quality\_Factor = PSNR + (100 * SSIM)
\end{equation}


\subsection{UAV-based Security System}
UAV-based public safety monitoring is a critical application which is often deployed in remote areas with limited bandwidth and charging facilities. Additionally, UAVs by design must deal with small battery life and limited storage space. Hence, this application operates in a space, power and bandwidth constraint environment. However, this application is resistant to imprecise storage and retrieval because it deals with images which retain most of the important features even after compression/approximation. And, a UAV roaming over different regions captures plenty of unnecessary images which may not be important for the application's goal/purpose. Hence there is a notion of object(s)-of-interest. All these characteristics and requirements, make this application ideal for using NS.


\subsubsection{Dataset Description}
To capture the application scenario, we have created a dataset containing UAV footage of a parking-lot \cite{ref:BINGO_DS}. The UAV remains in motion and captures images of cars and people in the parking lot. We construct the experiments with the assumption that the system user wishes to prioritize car images and perform frequent car image retrieval tasks.

\subsubsection{Effectiveness of NS in Comparison to Traditional CAM}
We notice a similar trend as observed for the wildlife surveillance system. 
The memory space utilization graph in Fig.~\ref{fig:CarRes}(a), shows that the NS framework is much more space-efficient than traditional CAM. In the zoomed-in portion, Fig.~\ref{fig:CarRes}(b), we observe that the memory space utilization decays after the store phase, due to compression of data which are not being accessed. In Fig.~\ref{fig:CarRes}(d-e), we observe that NS is more efficient in terms of retrieval operation cost due to its learning guided dynamic memory organization (operation cost is estimated as described in Section~\ref{bingoSimulator}). From Table~\ref{BINGO_collectedResTable}, we observe that NS is about 65x more efficient in terms of average operation cost (store and retrieve combined). Furthermore, in Table~\ref{BINGO_collectedResTable}, we note that the PSNR and SSIM of the fetched images during retrieve operations are similar for NS and traditional CAM. So it evident that the NS framework is equally effective as a traditional CAM in terms of serving the application. 



\subsubsection{Effectiveness of NS in Constrained Settings}
The UAV-based surveillance system may require to operate in resource-constrained environment. In Fig.~\ref{fig:CarRes}(c), we observe that the NS framework is much more suitable when it comes to functioning at extremely low memory space. On the other hand, the quality factor (defined in Eqn.~\ref{eqn1}) of the traditional CAM is much lower and increases very slowly as the storage space limitation is relaxed.

\section{Conclusion}
\label{conclusion}
We have presented NS, a learning-guided memory paradigm, which can provide a dramatic improvement in memory access performance and effective storage capacity in diverse applications. It draws inspiration from the human brain to systematically incorporate learning in the memory organization that dynamically adapts to the data access behaviour for improving storage and access efficiency. We have presented the retrieve, store and retention processes in details that integrate and employ data-driven knowledge. We have developed a complete performance simulator for NS and compared its data storage behaviour with traditional content-based memory. Quantitative evaluation of NS for two representative applications shows that it vastly surpasses the storage and retrieval efficiency of traditional CAM. By dynamically adapting data granularity and adjusting the associations between data and search patterns, NS demonstrates a high-level of plasticity that is not manifested by any existing computer memory organization. While we have worked with high-level memory organizational parameters here, our future work will focus on the physical implementation of NS. We believe, the proposed paradigm can open up avenues for promising physical realizations to further advance the effectiveness of learning and can significantly benefit from the data storage behaviour of emergent non-silicon nanoscale memory devices (such as resistive or phase change memory devices).

\bibliographystyle{IEEEtran}
\bibliography{IEEEabrv,chakra}


\vskip 0pt plus -1fil
\begin{IEEEbiography}[{\includegraphics[width=1in,height=1.2in,clip]{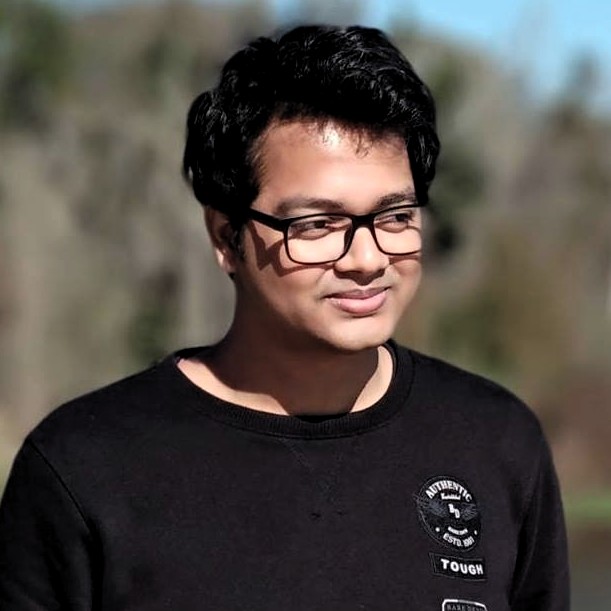}}]{Prabuddha Chakraborty} is pursuing his Ph.D. at the University of Florida under the supervision of Dr. Swarup Bhunia. He received his M.Tech. from Indian Institute of Technology (IIT), Kanpur. He has interned with Xilinx and Texas Instruments. His research interests include computer vision, system security and applied machine learning in various domains.
\end{IEEEbiography}
\vskip 0pt plus -1fil
\begin{IEEEbiography}[{\includegraphics[width=1in,height=1.2in,clip]{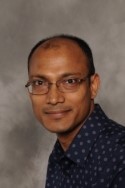}}]{Swarup Bhunia}
received his Ph.D. from Purdue University, IN, USA. Currently, Dr. Bhunia is a professor and Semmoto Endowed Chair of IoT in the University of Florida, FL, USA. Earlier he was appointed as the T. and A. Schroeder associate professor of Electrical Engineering and Computer Science at Case Western Reserve University, Cleveland, OH, USA. He has over ten years of research and development experience with over 200 publications in peer-reviewed journals and premier conferences. His research interests include hardware security and trust, adaptive nanocomputing and novel test methodologies. Dr. Bhunia received IBM Faculty Award (2013), National Science Foundation career development award (2011), Semiconductor Research Corporation Inventor Recognition Award (2009), and SRC technical excellence award as a team member (2005), and several best paper awards/nominations. He has been serving as an associate editor of IEEE Transactions on CAD, IEEE Transactions on Multi-Scale Computing Systems, ACM Journal of Emerging Technologies; served as guest editor of IEEE Design \& Test of Computers (2010, 2013) and IEEE Journal on Emerging and Selected Topics in Circuits and Systems (2014). He has served in the organizing and program committee of many IEEE/ACM conferences. He is a senior member of IEEE. 
\end{IEEEbiography}


\clearpage

\appendices

\section{Hyperparameters}
\label{hyperparameter_appendix}
In this appendix, we describe the NS hyperparameters in greter details. 
These hyperparameters can be set/changed by the user during the setup or during the operational stage. We define the \textbf{number of memory hives} as a system-level hyperparameter and we propose the following hyperparameters for each \textbf{memory hive}:
\begin{enumerate}
    \item \textbf{Number of localities ($L_n$)}: Each locality is used to store a specific nature of data. It is an unsigned integer value. If there exist $X$ types of objects-of-interest for an application, then using $X+1$ localities is advised. Every object-type can be assigned to a specific locality for optimal search efficiency and data granularity control. The last locality can be used for storing the unimportant data.
    \item \textbf{Memory decay rate of each locality}: Controls the rate at which data neuron memory strength and features are lost due to inactivity. It is a list (of length $L_n$) containing positive floating-point values. Assume that two localities $L1$ and $L2$ store separate object-types having importance $I1$ and $I2$ respectively. If $I1 > I2$, then it is advised to pick the decay rate of $L1$ to be less than the decay rate of $L2$.
    
    \item \textbf{Association decay rate of each locality}: Controls the rate at which NMN associations losses strength due to inactivity. It is a list (of length $L_n$) containing positive floating-point values. Assume that two localities $L1$ and $L2$ store separate object-types having importance $I1$ and $I2$ respectively. If $I1 > I2$, then it is advised to pick the decay rate of $L1$ to be less than the decay rate of $L2$.
    
    \item \textbf{Mapping between data features and localities}: This mapping dictates the segregation of application relevant data and their assignment to a locality with a lower decay rate. It is a dictionary with $L_n$ entries (one for each locality). Each entry is a list of data features (vectors) which when present in a data makes it a fit for the respective locality.
    \item \textbf{Data features and cue extraction AI (Artificial Intelligence) models}: These models are used to obtain more insights about the data during store/retrieve. They should be selected based on the data-type being processed. 
    
    \item \textbf{Data neuron matching metric}: Used during retrieve operation for finding a good match and during store operation for data neuron merging. For example, this metric can be something like \textbf{cosine similarity}.
    
    
    
    \item \textbf{Neural elasticity parameters}: Used to make space, in case the memory becomes full. It is a dictionary with $L_n$ entries. Each entry (corresponding to each locality) is a list of positive floating-point values. The values indicate the amount of memory strength loss imposed in successive iterations of the elasticity procedure.
    \item \textbf{Association weight adjustment parameter}: Used as a step size for increasing/decreasing association weights inside the NMN. A higher value will increase the dynamism but lower the stability. An optimal balance should be determined based on the application and based on a test-run.
    \item \textbf{Minimum association weight ($\varepsilon$)}: It is an unsigned integer which limits the decay of association weight beyond a certain point. A lower value will increase dynamism.
  
    \item \textbf{Degree of allowed impreciseness($\varphi$)}: Limits the amount of data feature which is allowed to be lost due to memory strength decay and inactivity. It is a floating-point number in the range of [ 0 - 100 ]. 0 implies data can get completely removed if needs arise. Keeping the parameter at 1 will ensure everything remains in the memory while retaining some unimportant memories at extremely low quality.
    
    \item \textbf{Frequency of retention procedure}: The retention procedure of NS brings in the effect of ageing. This hyperparameter is a positive integer denoting the number of normal operations to be performed before the retention procedure is called once. A lower value will increase dynamism at a cost of operation effort (energy and time consumption).
    
    \item \textbf{Compression techniques}: For each memory hive an algorithm for data compression must be specified. For example, we can use JPEG \cite{ref:jpeg} compression for an image hive.
    
\end{enumerate}
Additional hyperparameters can be added to the system to allow more fine-tuned adjustments. 

\section{Operation Algorithms}
\label{opAlgo_appendix}
In this appendix we will provide implementation level details of different NS operations, sub-operations and procedures.

\subsubsection{Write/Store Operation}

For storing a data element in the NS framework, we propose the algorithm depicted in Algo.~\ref{alg:store}. $M$ is the NS memory system, $D$ is the data to be stored, $S$ is the search parameters used for data merging attempts, $SearchLimit$ limits the amount of search effort spent on the merge attempt, $up$ can limit the number of changes made to the data neuron search order, and $k$ allows/disallows association weight decay. At first, the memory hive $MH$ suitable for the data type $DT$ (estimated from $D$) is selected (line 3). $Params$ is the set of hyperparameters associated with $MH$ (line 4). $\eta$ (step-size) is the association weight adjustment parameter (line 5). From $D$, the data features are extracted and stored in $DF$ (line 6). $S$ has three components: $C$ is the set of cues provided by the user that are associated with the given data $D$, $T1$ and $T2$ are used as the association and data feature matching thresholds respectively. In line 10, more cues are extracted (optionally) using AI models from $D$. The first step of data storage is to ensure that the memory hive $MH$ has enough space to store the incoming data. If the memory lacks space, then, in order to emulate a virtually infinite memory, the framework reduces data features and details of less important data until enough space is created. For example, in a wildlife surveillance system for detecting wolves, a lack of space can lead to compression of image frames without wolves in them. This is an iterative procedure (lines 13-15) and the data-reduction/compression aggressiveness increases with each iteration as shown in Algo.~\ref{alg:elasticity}. Once enough space is made for the incoming data $D$, the next step is to determine the locality $L$, which would be ideal for storing the data $D$ (line 16). $L$ is determined based on the hyperparameter which holds the mapping between data features and localities. Next, a search is carried out to determine if the data $D$ can be merged with an existing data neuron (lines 21-30). Intelligently merging the incoming data with a pre-existing data neuron can help save space and increase search efficiency. For example, if the incoming data is almost the same as an already existing data, then there is no need to separately store them. Before the search begins, the search order list is extracted and stored in $SO$ (using Algo.~\ref{alg:fetchSearchOrder}). The search order list is maintained as a part of the memory $M$ and is a dictionary containing $x$ entries where $x$ is the number of cues currently in the cue bank for the hive $MH$. Each entry is an ordered list of $<Path, DU>$ pairs arranged in the order of decreasing search priority. The search terminates when either a good match is found or when the $SearchLimit$ is reached.
During the search, at each iteration, if the data feature of the candidate/target data neuron ($TargetDN \rightarrow DF$) has a good match with the data feature of incoming data $DF$, then the $TargetDN$ is selected for merging. Using the $Reaction$ procedure (Algo.~\ref{alg:reaction}), the NMN is updated using reinforcement learning to reflect a success (line 26). Otherwise, the $Reaction$ procedure (Algo.~\ref{alg:reaction}) updates the NMN using reinforcement learning considering a failed merge attempt (line 28). If the merge candidate search terminates without finding a good match, then a new data neuron ($DN\_New$) is initialized inside the Locality $L$ (line 32) and all the cue-neurons corresponding to the respective cues in $C$ are associated (if already associated, then strengthened) with it (lines 33-34). If any of these cues are not present in the cue-bank, then new cue neurons are generated for those cues. The memory search order is also updated for $MH$ to reflect the presence of the new data neuron. These $<$CN, DN$>$ associations thus formed are also a form of learning.

The REACTION (Algo.~\ref{alg:reaction}) subroutine is a reinforcement learning guided procedure which is used during the store and retrieve operation for creating new associations, changing association weights, and updating search order for cues. $MH$ is the memory hive being updated, $TargetDN$ is the data neuron which is the focus of the change, $Path$ is the path to the $TargetDN$, $\eta$ (step-size) is the association weight adjustment parameter, $flag$ is used to indicate a search/merge success or failure, $C$ is the set of cues used during the search/merge procedure, $up$ is used to allow or disallow memory search order change, and $k$ allows/disallows association weight decay. Each association/connection $a$ in the $Path$, is either weakened or enhanced depending on the value of $flag$ and $k$, by an amount $\eta$ (lines 2-6). If $flag == 1$, then the memory strength of $TargetDN$ is increased (line 8). All the cues $C_i$ in $C$ are associated with $TargetDN$ and if a link exists, then it is strengthened (lines 9-10). Also, if any of these cues are not present in the cue-bank, then new cue neurons are generated for the cues. Finally, if $up == 1$, then the memory search order for the memory hive $MH$ is updated to reflect the alternations in NMN.

\begin{algorithm}[!t]

\scriptsize\addtolength{\tabcolsep}{-1.3pt}
\caption{Store}\label{alg:store}
\begin{spacing}{1.5}
\begin{algorithmic}[1]
\Procedure{store}{$M, D, S, SearchLimit, up, k$}
\State $DT = D \rightarrow Data\_Type$
\State $MH = select\_Hive(M, DT)$
\State $Params = MH \rightarrow Params$
\State $\eta = Params \rightarrow Assoc\_Wt\_Adj$
\State $DF = D \rightarrow Features$
\State $C = S \rightarrow Search\_Cues$
\State $T1 = S \rightarrow Assoc\_Thresh$
\State $T2 = S \rightarrow Match\_Thresh$
\State $C\_ext = extract\_Cue(Params, D)$
\State $C.append(C\_ext)$
\State $elasticity\_iter = 0$
\While{$size(D) > remaining\_space(MH)$}
    \State $ELASTICITY(elasticity\_iter, MH)$
    \State $elasticity\_iter++$
\EndWhile
\State $L = select\_Locality(Params, C, D)$
\State $found = False$
\State $SO = GET\_SEARCH\_ORDER(C,MH,T1,SearchLimit)$
\State $index = 0$
\State $DN\_List= \phi$
\While{$found == False$ \&\& $index <= SearchLimit$}
    \State $\{Path, TargetDN\} = SO[index]$
    \If{$TargetDN \notin DN\_List$}
        \If{$match(TargetDN \rightarrow DF, DF) > T2$}
            \State $found = True$
            \State $REACTION(MH, TargetDN, Path, \eta, 1, C, up, k)$
        \Else
            \State $REACTION(MH, TargetDN, Path, \eta, 0, C, up, k)$
        \EndIf
        \State $DN\_List.append(TargetDN)$
    \EndIf

    \State $index++$
    
\EndWhile

\If{$found == False$}
    \State $DN\_New =$ Initialize New DN with D in L
    \ForEach {$c \in \mathcal C$}
        \State $associate(c, DN\_New)$
    \EndFor
    \State $UPDATE\_MEMORY\_SEARCH\_ORDER(MH)$
\EndIf
\EndProcedure
\end{algorithmic}
\end{spacing}

\end{algorithm}
\subsubsection{Read/Load Operation}
For the retrieval/load operation, we propose the algorithm depicted in Algo.~\ref{alg:retrieve}. In the current implementation, this operation only returns one data-unit as output. $M$ is the NS memory system, $S$ is the search parameters used for retrieval, $SearchLimit$ limits the amount of search effort spent on the searching attempt, $up$ can limit the number of changes made to the search order, and $k$ allows/disallows association weight decay. At first the memory hive $MH$ is selected based on $DT$ in line 3 (the data type of $D$). $Params$ is the set of hyperparameters of $MH$ (line 4). From $S$ different components are extracted. $C$ is the set of search cues provided by the memory user, $DT$ is the data type, $T1$ and $T2$ are used as the association and data feature matching thresholds respectively. Additional cues are extracted optionally using AI models. $C1$ is the set of coarse-grained cues used for the NMN traversal and $C2$ is used for determining the data neuron with a good match. Before the search beings, the search order list is extracted and stored in $SO$ (using Algo.~\ref{alg:fetchSearchOrder}). Next, the search is carried out to determine if a good match can be found (lines 18-28).
During the search, at each iteration, if the data feature of $TargetDN$ has a good match with any of the fine-grained cues $C2$, then the $TargetDN$ is selected for retrieval. In this situation, the $Reaction$ procedure (Algo.~\ref{alg:reaction}), updates the NMN using reinforcement learning to reflect a success (line 23). Otherwise, the $Reaction$ procedure (Algo.~\ref{alg:reaction}) updates the NMN using reinforcement learning considering a failure (line 26). Finally, the data selected for retrieval ($Ret\_Data$) is returned.


\begin{algorithm}[!t]

\scriptsize\addtolength{\tabcolsep}{-1.3pt}
\caption{Retrieve}\label{alg:retrieve}
\begin{spacing}{1.5}
\begin{algorithmic}[1]
\Procedure{retrieve}{$M, S, SearchLimit, up,k$}
\State $DT = S \rightarrow Data\_Type$
\State $MH = select\_Hive(M, DT)$
\State $Params = MH \rightarrow Params$
\State $\eta = Params \rightarrow Assoc\_Wt\_Adj$
\State $C = S \rightarrow Search\_Cues$
\State $C\_ext = extract\_Cue(S \rightarrow Ref\_D)$
\State $C.append(C\_ext)$
\State $C1 = C \rightarrow Search\_Cues\_Coarse$
\State $C2 = S \rightarrow Search\_Cues\_Fine$
\State $T1 = S \rightarrow Assoc\_Thresh$
\State $T2 = S \rightarrow Match\_Thresh$

\State $found = False$
\State $DN\_List= \phi$
\State $Ret\_Data = \phi$
\State $SO = GET\_SEARCH\_ORDER(C1, MH, T1, SearchLimit)$
\State $index = 0$
\While{$found == False$ \&\& $index <= SearchLimit$}
    \State $\{Path, TargetDN\} = SO[index]$
    \If{$TargetDN \notin DN\_List$}
        \If{$match(TargetDN, C2) > T2$}
            \State $found = True$
            \State $REACTION(MH, TargetDN,Path, \eta, 1, C, up,k)$
            \State $Ret\_Data = TargetDN \rightarrow Data$
            
        \Else
            \State $REACTION(MH, TargetDN, Path, \eta, 0, C, up,k)$
        
        \EndIf
        \State $DN\_List.append(TargetDN)$
    \EndIf
    \State $index ++$
    
\EndWhile
\State \textbf{return}  $Ret\_Data$
\EndProcedure
\end{algorithmic}
\end{spacing}

\end{algorithm}

\subsubsection{Retention}
\begin{algorithm}[!t]

\scriptsize\addtolength{\tabcolsep}{-1.3pt}
\caption{Retention}\label{alg:retention}
\begin{spacing}{1.5}
\begin{algorithmic}[1]
\Procedure{retention}{$M, N, k$}
\State $C = M \rightarrow Connections$
\State $D = M \rightarrow Data\_Neurons$
\If{$k == 1$}
    \ForEach {$c \in \mathcal C$}
        \If{$Not\_Accessed\_Recently(c,N)$}
            \State $MH\_ID = Find\_Memory\_Hive\_ID(c)$
            \State $L\_ID = Find\_Memory\_Locality\_ID(c)$
            \State $a\_decay = Find\_Assoc\_Decay(MH\_ID, L\_ID, M)$ 
            \State$ Weaken(c, a\_decay)$
        \EndIf
    \EndFor
\EndIf

\ForEach {$d \in \mathcal D$}
    \If{$Not\_Accessed\_Recently(d,N)$}
        \State $MH\_ID = Find\_Memory\_Hive\_ID(d)$
        \State $L\_ID = Find\_Memory\_Locality\_ID(d)$
        \State $d\_decay = Find\_Mem\_Decay(MH\_ID, L\_ID, M)$ 
        \State $imp\_degree = Find\_Imprec\_Degree(MH\_ID)$
        \State $d\_str\_new = reduce\_mem\_str(d,d\_decay)$
        \State $Compress\_Mem(d, d\_str\_new, imp\_degree)$
    \EndIf
\EndFor
\ForEach {$MH \in \mathcal M$}
    \State $UPDATE\_MEMORY\_SEARCH\_ORDER(MH)$
\EndFor
\EndProcedure
\end{algorithmic}
\end{spacing}

\end{algorithm}
In the human brain, memories lose features and prominence over time. To model this sub-conscious process, we introduce the retention procedure. This procedure can be repeated after a particular interval or can be invoked after certain events. The reinforcement learning guided algorithm to carry out this operation is depicted in Algo.~\ref{alg:retention}. $M$ is the memory system, $N$ is the history threshold, and $k$ allows/disallows association weight decay. 
During this operation, any connections/associations not accessed during the last N-operations are weakened due to inactivity (lines 5-10) and any data neurons not accessed in the last N-operations are subject to feature-loss/compression (lines 11-18). The compression rate is limited by the maximum allowed degree of impreciseness ($imp\_degree$) for the given memory hive ($MH\_ID$). Also, the search order for the cues of each memory hive $MH$ is updated to reflect any changes due to alternation of association weights (lines 19-20).


\begin{algorithm}[!t]

\scriptsize\addtolength{\tabcolsep}{-1.3pt}
\caption{Reaction}\label{alg:reaction}
\begin{spacing}{1.5}
\begin{algorithmic}[1]
\Procedure{reaction}{$MH, TargetDN, Path, \eta, flag, C, up, k$}

\ForEach {$a \in \mathcal Path$}
    \If{$flag == 1$}
        \State $Increase\_Assoc\_Weight(a, \eta)$
    \EndIf
    \If{$flag == 0$ \&\& $k == 1$}
        \State $Decrease\_Assoc\_Weight(a, \eta)$
    \EndIf
\EndFor
    
\If{$flag == 1$}
    \State $Increase\_Mem\_Str(TargetDN)$
    \ForEach {$C_i \in \mathcal C$}
         \State $associate\_enhance(C_i,TargetDN)$
    \EndFor
\EndIf
\If{$up == 1$}
    \State $UPDATE\_MEMORY\_SEARCH\_ORDER(MH)$
\EndIf
\EndProcedure
\end{algorithmic}
\end{spacing}

\end{algorithm}

\subsection{Elasticity}
The ELASTICITY (Algo.~\ref{alg:elasticity}) subroutine is used during the store operation for making space in-case of a memory shortage. NS framework is designed to operate as a virtually infinite memory where no data is ever deleted but instead unimportant data are subject to feature loss over time. The elasticity hyperparameters are first extracted into $elast\_param$ from the memory hive $MH$ (line 2). Depending on the current iteration of elasticity ($elasticity\_iter$) and the Locality ($L$), we obtain the factor of memory strength decay (line 5), $ef$. For each data neuron $D$ in the locality $L$, the new memory strength is computed and are compressed if required (lines 6-8). The compression rate is limited by the maximum allowed degree of impreciseness ($imp\_degree$) for the given memory hive ($MH$).

\subsection{Get Search Order}
Algo.~\ref{alg:fetchSearchOrder} is a subroutine which is used to fetch the search order for a search/merge-attempt. $C$ is the set of cues provided for the operation, $MH$ is the memory hive where the search/merge-attempt is to take place, $T1$ is the average association weight threshold used to prune candidate data neurons and $SearchLimit$ is used to limit the number of candidates selected. For each cue $C_i$ in $C$, the search order list is fetched from $MH$ and stored in $SO$(line 5). Then for each candidate in $SO$, if the average association strength of the candidate is greater than $T1$, the $candidate$ is appended to the $CandidateList$. The final $CandidateList$ is returned at the end of the function (line 14) or at line 11, in case of an early exit.


\subsection{Update Memory Search Order}
Algo.~\ref{alg:updateMemorySearchOrder} is a subroutine which is used to update the search order for a given memory hive ($MH$). $Cues$ holds all the cues in $MH$. For every cue, $C$ in $Cues$, the paths to each data-neuron $D$, with the highest average association strength is selected and stored in decreasing order of average association strength. The $NewSearchOrder$ replaces the previous search order for the $MH$ (line 12).

\subsection{Update Operation}
In traditional CAM, update operation corresponds to changing the data associated with a particular tag/cue or by changing the tag/cue associated with a particular data. Such updates are also possible in NS using retrieve, store and retention procedures. For example, to associate a new tag/cue with a data, one can simply issue a retrieve operation of the target data with the new cue. This will cause the NS system to associate the new cue with the target data. Any old associations of the data with other cues will lose prominence over time if those associations do not get referenced. The storage, retrieval and retention algorithms can be used in many different ways to automatically form new associations and modify existing associations. Traditional CAM also supports data deletion which can be also enabled in NS by setting `degree of allowed impreciseness' hyperparameter to 0. Deletion is not directly achieved in NS but less important data (worthy of deletion) will slowly decay in-terms of memory strength due to lack of access and ultimately disappear. 

\begin{algorithm}[!t]

\scriptsize\addtolength{\tabcolsep}{-1.3pt}
\caption{Elasticity}\label{alg:elasticity}
\begin{spacing}{1.5}
\begin{algorithmic}[1]
\Procedure{elasticity}{$elasticity\_iter, MH$}

\State $elast\_param = MH \rightarrow Params \rightarrow elast\_param$
\State $imp\_degree = Find\_Imprec\_Degree(MH\_ID)$
\ForEach {$L \in \mathcal MH$}
    \State $ef = elast\_param[L \rightarrow Index][elasticity\_iter]$
    \ForEach {$D \in \mathcal L$}
        \State $d\_str\_new = Decrease\_Mem\_Str(D, ef)$
        \State $Compress\_Mem(D, d\_str\_new, imp\_degree)$
    \EndFor
\EndFor

\EndProcedure
\end{algorithmic}
\end{spacing}

\end{algorithm}
\begin{algorithm}[!t]

\scriptsize\addtolength{\tabcolsep}{-1.3pt}
\caption{Get Search Order}\label{alg:fetchSearchOrder}
\begin{spacing}{1.5}
\begin{algorithmic}[1]
\Procedure{GET\_SEARCH\_ORDER}{$C, MH, T1, SearchLimit$}
\State $CandidateList = \phi$
\State $Count = 0$
\ForEach {$C_i \in \mathcal C$}
    \State $SO = MH\rightarrow Search\_Order[C_i \rightarrow Index]$
    \ForEach {$candidate \in \mathcal SO$}
        \If{$candidate \rightarrow assoc\_Str > T1$}
            \State $CandidateList.append(candidate)$
            \State $Count++$
            \If{$Count >= SearchLimit$}
                \State \textbf{return}  $CandidateList$
            \EndIf
        \Else
            \State Break Loop
        \EndIf
        
    \EndFor
\EndFor
\State \textbf{return}  $CandidateList$


\EndProcedure
\end{algorithmic}
\end{spacing}

\end{algorithm}
\begin{algorithm}[!t]

\scriptsize\addtolength{\tabcolsep}{-1.3pt}
\caption{Update Memory Search Order}\label{alg:updateMemorySearchOrder}
\begin{spacing}{1.5}
\begin{algorithmic}[1]
\Procedure{UPDATE\_MEMORY\_SEARCH\_ORDER}{$MH$}
\State $Cues = MH \rightarrow Cues$
\State $NewSearchOrder = \phi$
\ForEach {$C \in \mathcal Cues$}
    \State $cueSearchOrder = \phi$
    \State $D =$ All data-neuron reachable from C
    \ForEach {$D_i \in \mathcal D$}
        \State $SP = find\_Average\_Strongest\_path(C, D_i)$
        \State $assoc\_Str = average\_Assoc\_Str(SP)$
        \State $insert\_Sort(cueSearchOrder, {DU,SP,assoc\_Str})$
    \EndFor
    \State $NewSearchOrder[C \rightarrow Index] = cueSearchOrder$
\EndFor
\State $MH \rightarrow Search\_Order = NewSearchOrder$
\EndProcedure
\end{algorithmic}
\end{spacing}

\end{algorithm}

\section{NS Simulator Case-Study Configuration}
\label{expt_details_appendix}

In this appendix, we shall provide the NS simulator configuration used during the case-studies. In the context of the case studies, we allow the following simplifications over the proposed organization and algorithms (presented in Appendix~\ref{opAlgo_appendix}):
\begin{itemize}
    \item Connections between cue-neurons are not formed.
    \item Search order entries have a path length of 1.
    \item Every locality has a ''default cue" which is connected with all the data-neurons in the locality and is used to access data-neuron which are not otherwise accessible from normal cues. This construct emulates $<DN, DN>$ associations.
\end{itemize}


\subsection{NS hyperparameters Used}
For all the case studies, we use a single memory hive for holding the image data. For this specific hive we use the following hyperparameters:

\begin{itemize}
    \item Number of localities ($L_n$): 2.
    
    \item Memory decay rate of each locality: [0.5, 1]
    
    \item Association decay rate of each locality: [0, 0]
    
    \item Mapping between data features and localities: Depends on the application and case-study. 
    \begin{itemize}
    \item \textbf{Wildlife Surveillance:} For scenario 1, deer images are mapped to locality 0 and other images are mapped to locality 1. For scenario 2, wolf/fox images are mapped to locality 0 and other images are mapped to locality 1.
    \item \textbf{UAV-based Security System:} Car images are mapped to locality 0 and other images are mapped to locality 1.
    \end{itemize}
    \item Data features and cue extraction AI (Artificial Intelligence) models: VGG 16 predicted classes are used as coarse-grained cues \cite{vgg16}. For data feature and fine-grained cues, we use the VGG 16 fc2 activations.
    
    \item Data-Neuron matching metric: Cosine-Similarity.
    
    \item Neural elasticity parameters:
    
      0 $\rightarrow$ [80, 70, 60, 50, 40, 30, 20, 10, 1] \\
      1 $\rightarrow$ [80, 70, 60, 50, 40, 30, 20, 10, 1]

    \item Association weight adjustment parameter: 20

    \item Degree of allowed impreciseness: 1
    
    \item Frequency of retention procedure: 500
    
    \item Compression technique: We use JPEG compression for the image hive.

\end{itemize}

\subsection{NS Operation Parameters Used}
These operation parameters (as described in Appendix~\ref{opAlgo_appendix}) are used during all store, retrieve and retention operations/procedures unless otherwise specified. 
\begin{itemize}
    \item $S \rightarrow Assoc\_Thresh$: 0
    \item $S \rightarrow Match\_Thresh$: 0.95
    \item $SearchLimit$: Maximum (Int\_Max)
    \item $up$ = 1
    \item $k$ = 0
    
\end{itemize}

\begin{figure*}[h!]
\centering
\includegraphics[width=\textwidth]{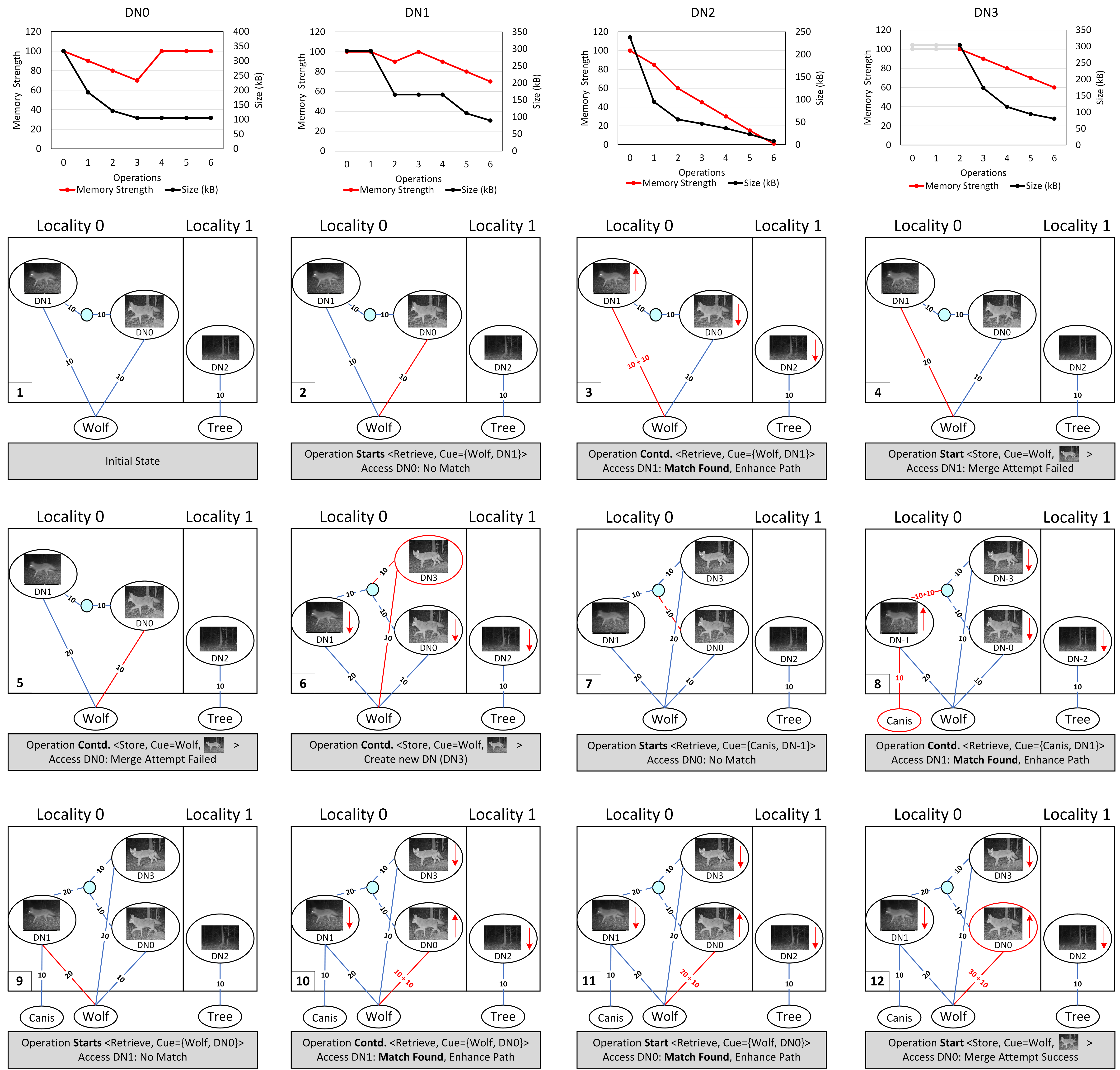}
\caption{A peek inside the NS framework for understanding its dynamism and plasticity.}
\label{fig:dynamism}
\end{figure*}
\section{Dynamism of NS: A Detailed Analysis}
\label{dynamism_appendix}
To visually illustrate how the dynamism of NS we showcase a series of NS framework snapshots in Fig.~\ref{fig:dynamism}. Same operation parameters and hyperparameters as mentioned in Appendix~\ref{expt_details_appendix} are used with the followings exceptions:
\begin{itemize}
    \item Memory decay rate of each locality: [10, 20]
    
    \item Mapping between data features and localities: All fox/wolf images are designated to stay in locality 0 and the rest are assigned to locality 1.
   
    \item Association weight adjustment parameter: 10

    \item Frequency of retention procedure: 1
\end{itemize}
The top 4 graphs in Fig.~\ref{fig:dynamism} plots the data neuron memory strength (Y-axis left) and size (Y-axis right) with respect to the number of operations performed (X-axis) for each of the 4 data neurons in the scenario. The 12 snapshots (ID provided on the lower-left corner) in Fig.~\ref{fig:dynamism} are described below:
\begin{enumerate}
    \item In the initial state, there are two data neurons in locality 0 and one data neuron in locality 1.
    
    \item The first operation starts with the coarse-grained cue ``Wolf" and the fine-grained cue corresponding to the data-feature of DN1. Using the coarse-grained cue ``Wolf", the NS framework first compares the fine-grained cue with the data-feature of DN0. The matching fails due to lack of similarity.
    
    \item In this step, the previous operation continues and the NS framework reaches DN1 using the coarse-grained cue ``Wolf". The fine-grained cue and data feature of DN1 matches leading to a successful retrieval. The weight of $<$Wolf, DN1$>$ association is increased. Also, note that the memory strength of all data neurons decreases and memory strength of DN1 is restored back to 100 (indicated by the red arrows).
    
    \item The next operation is of type store. The cue provided is ``Wolf" and a new data with nothing similar in the locality 0 is provided. The NS framework first attempts to merge the incoming data with an exiting data neuron. The first match with DN1 fails because they are not similar. Note that DN1 is searched first because $<$Wolf,DN1$>$ is greater than $<$Wolf,DN0$>$.
    
    \item The match with DN0 also fails due to lack of data similarity. 
    
    \item Given that the merge attempt failed, a new data neuron DN3 is generated for the new incoming data. DN3 gets associated with the other DNs via the default cue neuron (cyan coloured circle in middle) and also gets associated with cue ``Wolf". Furthermore, the memory strength of all remaining data neurons decreases.
    
    \item The next operation is a retrieve operation with coarse-grained cue ``Canis" and the fine-grained cue same as the data feature of DN1. There are no cue neuron for ``Canis", so the search is carried out via the default cue neuron (Locality-0). The first search yields DN0 which is not a good match.

    \item The second search yields the correct output. The $<$default cue neuron, DN1$>$ strength increase. A new cue neuron for ``Canis" is generated and linked with DN1. Also, the memory strength of all data neurons except DN1 decreases and memory strength of DN1 is restored back to 100.
    
    \item The next operation is also a retrieve operation with coarse-grained cue ``Wolf" and fine-grained cue same as data-feature of DN0. $<$DN1, Wolf$>$ has the highest strength among other ``Wolf" associations, so DN1 is compared with first. This however does not yield a good match.
    
    \item The next data-neuron searched is DN0 and a good similarity is found. The association $<$Wolf, DN0$>$ increases in strength. Memory strength of DN0 is restored back to 100 and the memory strength of all remaining data neurons decreases.
    
    \item The next operation is of type retrieve with coarse-grained cue ``Wolf" and fine-grained cue same as the data-feature of DN0. The association $<$Wolf, DN0$>$ is explored first (because being first in the search order for cue ``wolf") and a good match is found. Memory strength of DN0 is restored back to 100 and the memory strengths of all remaining data neurons decreases. Also, the $<$Wolf, DN0$>$ association is strengthened.
    
    \item The next operation is a store operation and a data very similar to DN0 is provided with cue ``Wolf". The first merge attempt with DN0 is a success and no new data neuron is generated for the incoming data. Memory strength of DN0 is restored back to 100 and the memory strengths of all remaining data neurons decreases. Also, the $<$Wolf, DN0$>$ association is strengthened.
    
\end{enumerate}

In the DN0 graph (Fig.~\ref{fig:dynamism}), we observe that the data size and memory strength is maintained at a relatively high value throughout the case-study. This is because DN0 have been accessed the most among all data neurons. For DN2 (background image in locality 1), the memory strength constantly decays at a higher rate due to lack of access and importance.

\section{Additional Applications Suitable for NS}
\label{addApp}

We have observed that the wildlife surveillance system and the UAV-based security system can benefit greatly from using NS. We believe that there are many other such applications suitable for NS which are widely being used in different domains. Some of these applications are discussed next.

\subsection{Agriculture Automation}
Automation in agriculture such as weed detection \cite{haug2014crop_weeddetection}, aerial phenotyping \cite{bauer2019combining_arielPhenotyping} and agriculture pattern analysis \cite{chiu2020agriculture_agriculturePattern} requires storing and subsequent analysis of huge amount of image data. A dynamic, intelligent and virtually infinite memory framework like NS can adapt to the demand while providing optimal performance. 

\subsection{Forest Fire Detection}
Forest fire detection and prompt reaction can save lives and reduce the impact on air/soil. Different proposed forest fire detection systems capture, store and analyse multi-modal data such as image, sound, gas profile, temperature, humidity etc. \cite{arialFireDetection, ref:firedrone, forestFireSensorNetwork}. Efficiently managing and segregating (important from unimportant) this huge amount of constantly streaming data is crucial for success and an intelligent priority-driven memory system such as NS can help optimize storage, retrieval and analysis performance. 

\subsection{Post-Disaster Infrastructure Inspection}
Automatic detection techniques for inspecting infrastructure damages due to natural disasters such as earthquake and hurricane are being widely explored \cite{UAVRoofHole_IoT, UAV_earthquake_IoT, ref:crackDS}. Most of these techniques deal with a huge influx of data that may not be always relevant to the task. An intelligent memory framework such as NS can optimize the entire data retention process to boost overall system performance. 

\subsection{Maritime Surveillance}
Detecting and monitoring maritime activities often involve SAR (Synthetic-aperture radar) data, standard radar data, infrared data and video data \cite{carthel2007multisensor_maritimeSurv}. Efficiently handling this huge amount of multi-modal data is crucial for success and NS would be ideal for such applications.

\subsection{Space and Remote Planet Exploration}
Space observation facilities \cite{way2012advances_dataminingForAstronomy} and planet exploration systems (such as the MARS Rover) \cite{carr2007surface_marsRoverBOok, rothrock2016spoc_MARsRover1} deal with huge influx of data. With limited space, energy and bandwidth constraints it is crucial to store and transmit efficiently. A framework like NS can certainly boost the efficiency of such a system.

\end{document}